\documentclass[journal]{IEEEtai}
\usepackage[colorlinks,urlcolor=blue,linkcolor=blue,citecolor=blue]{hyperref}

\usepackage{color,array}

\usepackage{graphicx}

\usepackage{times}

\usepackage{algorithm}
\usepackage{algpseudocode}

\usepackage[numbers]{natbib}

\usepackage{hyperref}
\usepackage{url}

\usepackage{booktabs}       % professional-quality tables
\usepackage{amsfonts}       % blackboard math symbols
\usepackage{nicefrac}       % compact symbols for 1/2, etc.
\usepackage{microtype}      % microtypography
\usepackage{colortbl}
\usepackage{xcolor}         % colors
\usepackage{amsmath}
\usepackage{graphicx}
\usepackage{pifont}

\usepackage[normalem]{ulem} 
\usepackage{multirow}
\usepackage{multicol}
\usepackage{subcaption}
\usepackage{wrapfig}

\usepackage[scr=boondox,cal=esstix]{mathalpha}
\usepackage{overpic}
\usepackage{algorithmicx}
\usepackage{listings}
\usepackage{etoolbox} 
\usepackage[final]{changes}
\makeatletter
\patchcmd{\Changes@output}{\IfIsEmpty{#3}}{\ifblank{#3}}{}{}
\makeatother

\usepackage{xr}
\usepackage{xcite}
\externalcitedocument{supplementary}

\usepackage{amsmath,amsfonts,bm}

\def\eqref#1{equation~\ref{#1}}
\def\1{\bm{1}}

\def\vy{{\bm{y}}}

\def\evy{{y}}

\DeclareMathAlphabet{\mathsfit}{\encodingdefault}{\sfdefault}{m}{sl}
\SetMathAlphabet{\mathsfit}{bold}{\encodingdefault}{\sfdefault}{bx}{n}

\def\gA{{\mathcal{A}}}

\def\gD{{\mathcal{D}}}

\def\gS{{\mathcal{S}}}
\def\gT{{\mathcal{T}}}

\def\gY{{\mathcal{Y}}}
\def\gZ{{\mathcal{Z}}}

\newcommand{\mypapername}{Future of Time Series Anomaly Detection: From Historical Review to Unified Outlook}
\newcommand{\anomURL}{\url{https://emorzz1g.github.io/FTSAD/}}

\newcommand{\mycolor}{black}

\newenvironment{parcolor}[1][\mycolor]
{\color{#1}}
{}

\begin{document}

\title{\mypapername}

\author{Zhijie~Zhong,~Zhiwen~Yu\(^*\),~\IEEEmembership{Senior~Member,~IEEE},~Pengyu~Li,~Liujian~Zhang,~Qiaoli~Ni,\\
Jiang~Jun,~Jianming~Lv,~\IEEEmembership{Member,~IEEE},~Kaixiang~Yang,~\IEEEmembership{Member,~IEEE},~and~C.~L.~Philip~Chen,~\IEEEmembership{Fellow,~IEEE}

\IEEEcompsocitemizethanks{
\IEEEcompsocthanksitem This work was supported in part by National Natural Science Foundation of China No. 92467109, U21A20478, National Key R\&D Program of China 2023YFA1011601, and the Major Key Project of PCL, China under Grant PCL2025A11 and PCL 2025A13.
\IEEEcompsocthanksitem Zhijie~Zhong and Liujian~Zhang are with the School of Future Technology, South China University of Technology, Guangzhou, 510650, China, and also with the Pengcheng Laboratory, Shenzhen, 518066, China.
\IEEEcompsocthanksitem Zhiwen~Yu is with the School of Computer Science and Engineering, South China University of Technology, Guangzhou, Guangdong 510650, China, and also with the Pengcheng Laboratory, Shenzhen, Guangdong 518066, China (\(^*\)denotes the corresponding author, Email: zhwyu@scut.edu.cn).
\IEEEcompsocthanksitem Qiaoli~Ni is with HUAWEI Technologies Co., Ltd., Shenzhen, Guangdong 518066, China.
\IEEEcompsocthanksitem Jianming~Lv is with the Institute of Computing Technology Chinese Academy of Sciences, South China University of Technology, Guangzhou, Guangdong 510641, China.
\IEEEcompsocthanksitem Jiang~Jun is with the Pengcheng Laboratory, Shenzhen, Guangdong 518066, China.
\IEEEcompsocthanksitem Pengyu~Li, Kaixiang~Yang and C.~L.~Philip~Chen are with the School of Computer Science and Engineering, South China University of Technology, Guangzhou 510650, China.
}}

\markboth{Journal of IEEE Transactions on Artificial Intelligence, Vol. 00, No. 0, Month 2020}
{Zhong \MakeLowercase{\textit{et al.}}: \mypapername}

\maketitle

\begin{abstract}
Time series anomaly detection (TSAD) is a foundational problem in data mining, aiming to identify observations that deviate significantly from nominal temporal patterns. Driven by escalating data volumes and application complexity, recent research has yielded diverse architectures, including diffusion models, state space models, and large language models (LLMs). However, existing surveys provide limited coverage of these emerging designs and lack a unified treatment of multivariate anomalies, cross-domain unified modeling, metric reliability, and contemporary evaluation challenges.
To bridge these gaps, this survey delivers three core contributions:
(1) We establish a refined anomaly categorization and a structured taxonomy of paradigms reflecting modern advances, comparing detection mechanisms and charting developmental trajectories across paradigms.
(2) We provide an extensive analysis of evaluation methodologies and conduct comprehensive experiments to assess metric fidelity, complemented by an open-source benchmark repository encompassing algorithms, datasets, and evaluation tools.
(3) We outline key open challenges, offering forward-looking perspectives on dynamic thresholding, model interpretability, zero-shot transfer, multi-modal alignment, and foundational models for time series.
By addressing existing theoretical and empirical limitations, this survey establishes a coherent framework to guide future TSAD research and practical deployments.
\end{abstract}

\begin{IEEEImpStatement}
Detecting anomalies in time series data is essential for safeguarding critical infrastructure across healthcare, finance, transportation, and manufacturing. Failures in detection can result in undetected equipment malfunctions, financial fraud, and security breaches, carrying significant economic and societal consequences. Despite rapid algorithmic progress, practitioners face a fragmented research landscape with inconsistent taxonomies and unreliable evaluation standards, making dependable method selection difficult. This survey addresses this challenge by establishing a unified paradigm taxonomy that separates detection principles from model architectures, empirically exposing the inflation bias of widely adopted metrics, and providing actionable algorithm selection guidelines tailored to operational constraints such as interpretability, latency, and zero-shot capability. By further charting open frontiers in dynamic thresholding, model interpretability, multi-modal perception, and foundational models, this work offers a structured roadmap to advance both research and real-world deployment of time series anomaly detection.
\end{IEEEImpStatement}


% \begin{table}[htbp]
%   \centering
%   \caption{List of Abbreviations}
%   \vspace{3mm}
% % \begin{multicols}{2}
% \begin{tabular}{l|l}
% \toprule
% \textbf{Abbreviation} & \textbf{Expansion} \\
% \midrule
% TSAD & Time Series Anomaly Detection \\
% SSM & State Space Model \\
% LLM & Large Language Model \\
% RNN & Recurrent Neural Network \\
% GNN & Graph Neural Network \\
% TCN & Temporal Convolution Network \\
% AE & Autoencoder \\
% VAE & Variational Autoencoder \\
% Prot & Prototype-based \\
% % \bottomrule
% % \end{tabular}

% % \begin{tabular}{l|l}
% % \toprule
% % \textbf{Abbreviation} & \textbf{Expansion} \\
% % \midrule
% Diff & Diffusion-based \\
% CC & Channel-correlation \\
% FFN & Feed-Forward neural Network \\
% Rec & Reconstruction \\
% Fcst & Forecasting \\
% Gen & Generative \\
% Sim & Similarity \\
% Cls & Classification \\
% \bottomrule
% \end{tabular}
%  \label{tab:abbr}%
% % \end{multicols}
% \end{table}%

\section{Introduction}
\IEEEPARstart{A}{} time series is a sequence of data points indexed in chronological order, with extensive applications across diverse domains, including finance, healthcare, industrial monitoring, and cybersecurity \cite{sv_choiDeepLearningAnomaly2021,sv_alimohammadiPerformanceEvaluationOutlier2022,app_ECGGAN2023,jj1}. Time series anomaly detection (TSAD), a fundamental area of temporal data analytics, aims to identify observations or segments that deviate significantly from nominal patterns \cite{cont_zhong2024simad,sv_deep2024, trs_yangDCdetectorDualAttention2023}. \added{By uncovering early indicators of equipment malfunctions, financial fraud, and malicious network intrusions, timely anomaly detection safeguards system reliability and prevents catastrophic operational failures across complex infrastructures \cite{app_swat,jj2,app_NAB_tail-p_threshold}.}

\begin{figure}[htbp]
    \centering
    \includegraphics[width=1\linewidth]{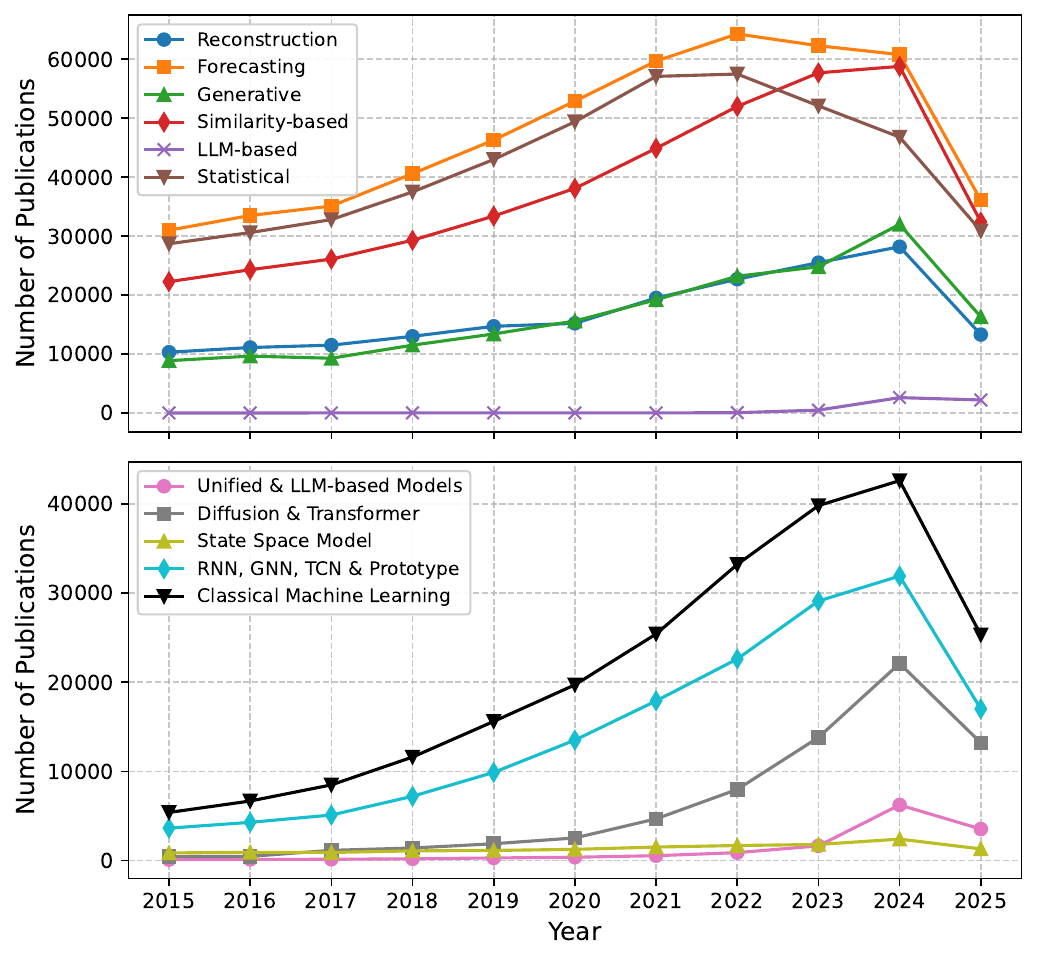}
    \caption{Publication trends of TSAD algorithms from 2015 to 2025. The top panel shows a statistical breakdown based on the taxonomy of algorithmic paradigms, while the bottom panel delineates publications across different model architectures.}
    \label{fig:trend}
\end{figure}

The field of TSAD has evolved rapidly, transitioning from early statistical and classical machine learning methods to advanced deep learning architectures. To quantitatively contextualize this evolution, we analyzed publication trends over the decade spanning 2015 to 2025 across different algorithmic paradigms and architectures using data from Google Scholar, \added{as illustrated in Fig.~\ref{fig:trend}}. \added{This analysis reveals several key shifts: while research into traditional forecasting and statistical paradigms has gradually decelerated since 2022, generative, similarity-based, and reconstruction paradigms remain highly active. Concurrently, large language model (LLM)-based methods have exhibited a marked surge in publication volume since 2024, emerging as the fastest-expanding research area. Detailed search queries and raw publication data are documented in Section~\ref{app:Related_Works} of the Supplementary Material (SM).}

However, this rapid expansion has also exposed notable limitations in existing surveys. As detailed in our related work comparison, previous surveys typically focus on specific application domains or earlier model families, leaving a marked gap in the coverage of recent architectural innovations, such as Transformers \cite{trs_yangDCdetectorDualAttention2023,trs_wangDisentangledDynamicDeviation2023,cont_zhong2024simad}, Diffusion Models \cite{dif_chenImDiffusionImputedDiffusion2023a,dif_huUnsupervisedAnomalyDetection2024}, State Space Models (SSMs) \cite{s6_ahamedTimeMachineTimeSeries2024,s6_ASSM,s6_MemMambaAD}, and Large Language Models (LLMs) \cite{sv_zhangLargeLanguageModels2024,llm_aLLM4TS2024,llm_AnomalyLLM2024,llm_GPT4TS2023}. In parallel, existing surveys lack a unified taxonomy to categorize these emerging paradigms, while critical methodological issues, such as anomaly classification for multivariate series and the empirical reliability of evaluation metrics, remain insufficiently examined. In particular, ambiguous anomaly definitions in multivariate time series continue to hinder reproducible research progress.

\begin{figure*}
    \centering
    \includegraphics[width=0.9\linewidth]{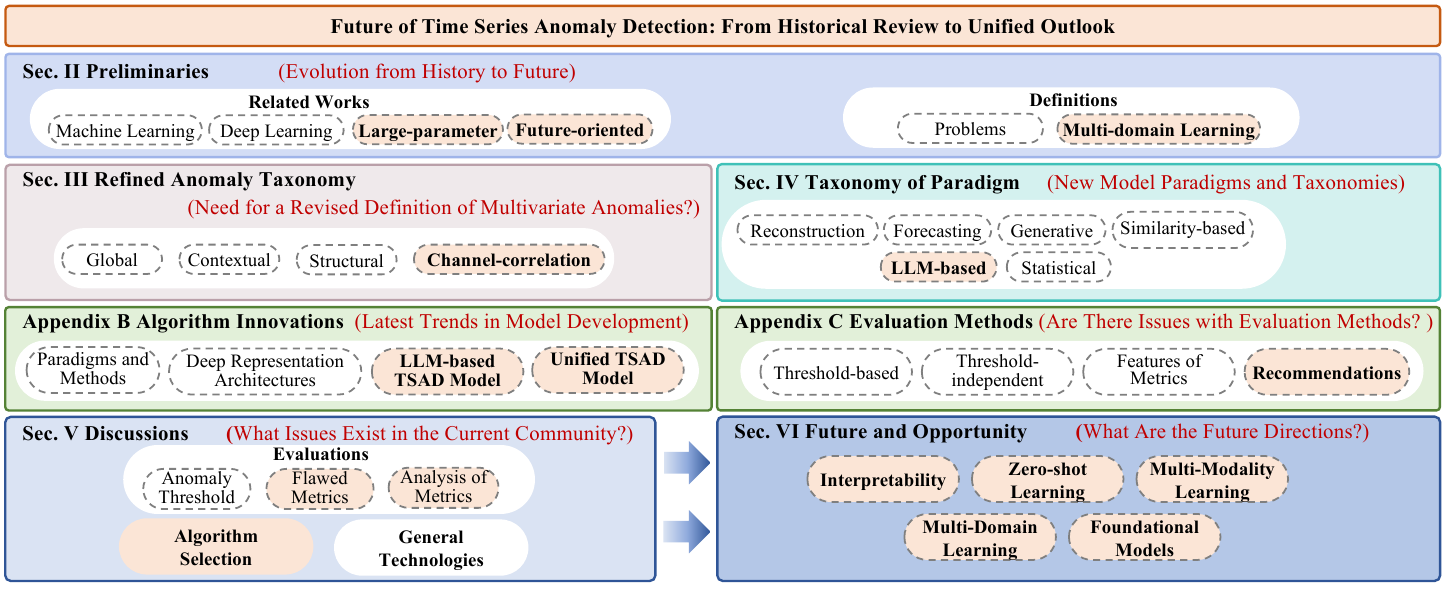}
    \caption{The organizational framework of this paper. }
    \label{fig:content}
\end{figure*}

To address these limitations, this survey presents a comprehensive and unified perspective on TSAD. Specifically, we introduce a refined classification for time series anomalies, establish a structured taxonomy of algorithmic paradigms grounded in core detection principles, and thoroughly examine representative deep learning architectures. In addition, we discuss community-wide challenges and outline promising future directions to guide subsequent research.
The organizational framework of this paper is illustrated in Fig. \ref{fig:content}.

In summary, our key contributions are:
\begin{enumerate}
    \item \textbf{Refined Anomaly Taxonomy} (Sec.~\ref{sec:Refined_Anomaly_Taxonomy}): We formalize the concept of channel-correlation anomalies, establishing a unified perspective on multivariate anomaly patterns and providing a rigorous framework for analyzing complex cross-channel dependencies.
    \item \textbf{Structured Taxonomy of Paradigms} (Sec.~\ref{sec:Taxonomy_of_Paradigm}): We establish a structured taxonomy that categorizes mainstream algorithms according to their core detection assumptions and structural characteristics, clarifying the fundamental mechanisms governing their detection capabilities.
    \item \textbf{Comprehensive Review of Emerging Architectures} (SM~\ref{sec:Algorithm_Innovations}): We provide an in-depth overview of modern algorithm designs, with dedicated analyses of diffusion models, state space models, prototype-based networks, and LLM-driven detectors that are largely omitted from previous surveys.
    \item \textbf{Critical Evaluation of Metrics and Benchmark Toolkit} (SM~\ref{sec:Evaluation_Methods} and Sec.~\ref{sec:Discussions}): We critically evaluate existing evaluation metrics, exposing methodological pitfalls such as metric inflation bias and offering actionable selection guidelines. To facilitate reproducible research, we compile and publicly release an open-source benchmark repository covering representative algorithms, datasets, and standardized evaluation tools.
    \item \textbf{Open Challenges and Future Directions} (Secs.~\ref{sec:Discussions} and \ref{sec:Future_and_Opportunity}): We highlight pivotal open challenges, including dynamic thresholding, model interpretability, zero-shot transfer, multi-modal alignment, and foundational models, outlining a structured roadmap for future research and deployment.
\end{enumerate}
\added{The remainder of this survey is structured as follows. Sec.~\ref{sec:Preliminaries} reviews related literature and defines essential terminology. Sec.~\ref{sec:Refined_Anomaly_Taxonomy} presents our refined anomaly taxonomy, followed by the structured taxonomy of algorithmic paradigms in Sec.~\ref{sec:Taxonomy_of_Paradigm}. Secs.~\ref{sec:Discussions} and \ref{sec:Future_and_Opportunity} provide in-depth discussions on operational challenges and outline prospective research opportunities. Finally, Sec.~\ref{sec:Conclusion} concludes the survey.}

\added{To foster reproducible research and community engagement, all resources mentioned in this survey, including supplemental materials, benchmark code, and curated datasets, are openly accessible at \anomURL.}

\section{Preliminaries}\label{sec:Preliminaries}
This section positions our survey within the broader context of TSAD research through a comparative analysis of related literature. We then formalize core mathematical notations and problem definitions, establishing the core concepts required for the anomaly taxonomy in Sec.~\ref{sec:Refined_Anomaly_Taxonomy} and the paradigm classification in Sec.~\ref{sec:Taxonomy_of_Paradigm}.

\subsection{Related Works}\label{sec:Related_Works}
To delineate the historical and technical trajectory of the field, we categorize the evolution of TSAD algorithms into four chronological phases governed by dominant architectural and modeling principles.

The \textbf{first phase} focused on \textit{classical machine learning algorithms}. During this period, general-purpose anomaly detectors (such as tree-based ensembles and empirical distribution estimators) were either applied directly to temporal sequences or augmented with handcrafted features \cite{ml_ECOD2022,ml_IsolationForest2008,jj3}. Although computationally lightweight and inherently interpretable, these methods frequently struggle with the scale and high dimensionality of modern datasets, as their underlying assumptions of linearity or low-rank structures fail to capture complex, nonlinear, and dynamic temporal dependencies.

The \textbf{second phase} introduced \textit{lightweight deep learning models for TSAD}. These architectures primarily encompass three families: Recurrent Neural Networks (RNNs) \cite{vae_suRobustAnomalyDetection2019,ae_zhouContrastiveAutoencoderAnomaly2022b}, Temporal Convolutional Networks (TCNs) \cite{tcn_abilashaWarpingResilientScalable2022,tcn_huangEfficientTimeSeries2022}, and Graph Neural Networks (GNNs) \cite{gcn_yangGraphTransformerNetwork2024,gcn_heUnidirectionalGraphStructure2024,gcn_huangMultivariateTimeSeries2024}. Concurrently, prototype-based models \cite{trans_MEMTO_2023,ae_zhangAdaptiveMemoryNetworks2022,AdaMemBLS} emerged as a distinct subcategory within the similarity-based paradigm. Unlike standard autoencoding architectures that frequently suffer from over-generalization, such as inadvertently reconstructing anomalous inputs, prototype models explicitly learn a discrete dictionary of representative memory prototypes to encapsulate nominal temporal dynamics. Because existing surveys have largely overlooked prototype-based algorithms, we thoroughly analyze their formulations and architectural characteristics.

The \textbf{third phase} centers on \textit{large-parameter deep learning models for TSAD}, characterized by expressive backbone architectures designed to capture extended temporal contexts. Transformer-based models exploit multi-head self-attention to model long-range sequence correlations \cite{trs_yangDCdetectorDualAttention2023,trs_wangDisentangledDynamicDeviation2023}, whereas diffusion-based models estimate underlying data distributions through iterative denoising to isolate low-probability events \cite{dif_huUnsupervisedAnomalyDetection2024,dif_pintilieTimeSeriesAnomaly2023}. More recently, state space models (SSMs) \cite{s6_ahamedTimeMachineTimeSeries2024,s6_MemMambaAD}, particularly selective scan architectures such as Mamba, achieve linear computational complexity with respect to sequence length, demonstrating substantial potential for scalable sequence analysis. Given that prior surveys have focused predominantly on Transformers while giving insufficient treatment to diffusion models and SSMs, we review both categories in detail to address this gap.

Finally, the \textbf{fourth phase} explores \textit{future-oriented deep TSAD algorithms}, primarily encompassing cross-domain unified architectures and large language model (LLM)-adapted detectors. Rather than training isolated models from scratch for each dataset, current research emphasizes zero-shot transfer, multi-modal signal alignment (integrating numeric sequences with textual logs and visual indicators), and foundational models for time series. We examine these emerging paradigms and their operational constraints in Secs.~\ref{sec:Discussions} and \ref{sec:Future_and_Opportunity}. Because existing surveys rarely explore LLM-based approaches or unified multi-domain architectures, our survey provides a timely roadmap for these frontiers.

\begin{table*}[ht]
    \caption{Overview of Recent Key Time Series Anomaly Detection Surveys.}
    \centering
    \resizebox{0.95\linewidth}{!}{
    \begin{tabular}{|c|c|c|c|cc|c|c|c|ccc|}
\toprule
\multicolumn{1}{|c|}{\multirow{2}[4]{*}{Citation}} & 
\multicolumn{1}{c|}{\multirow{2}[4]{*}{Application}} & 
\multicolumn{1}{c|}{\multirow{2}[4]{*}{\parbox{16mm}{\centering Taxonomy of Paradigms}}} & 
\multicolumn{1}{c|}{\multirow{2}[4]{*}{\parbox{15mm}{\centering Anomaly Taxonomy}}} & 
\multicolumn{2}{c|}{Novel Algorithms} & 
\multicolumn{1}{c|}{\multirow{2}[4]{*}{Library}} & 
\multicolumn{1}{c|}{\multirow{2}[4]{*}{Datasets}} & 
\multicolumn{1}{c|}{\multirow{2}[4]{*}{Metrics}} & 
\multicolumn{3}{c|}{Discussion} \\
\cmidrule(lr){5-6}\cmidrule(lr){10-12}   
&   &    &      & Prot/SSM/Diff & LLM &  &  &  & Interpretability & Multi-Modality & Multi-Domain \\
\midrule
\rowcolor[rgb]{0.851, 0.851, 0.851} \citet{sv_choiDeepLearningAnomaly2021} (2021) & \ding{51} & \ding{51} &    &    &    &    & \ding{51} &    &    &    &  \\
\citet{sv_schmidlAnomalyDetectionTime2022} (2022) &    & \ding{51} & \ding{51} &    &    &    & \ding{51} & \ding{51} &    &    &  \\
\rowcolor[rgb]{0.851, 0.851, 0.851} \citet{sv_wenTransformersTimeSeries2023} (2023) &    &    &    &    &    &    &    &    &    &    & \ding{51} \\
\citet{sv_yanComprehensiveSurveyDeep2024} (2023) & \ding{51} & \ding{51} & \ding{51} &    &    &    &    &    &    &    &  \\
\rowcolor[rgb]{0.851, 0.851, 0.851} \citet{sv_yeSurveyTimeSeries2024} (2024) & \ding{51} &    &    &    & \ding{51} &    &    &    & \ding{51} & \ding{51} & \ding{51} \\
\citet{sv_su2024large} (2024) & \ding{51} &    &    &    & \ding{51} &    & \ding{51} & \ding{51} & \ding{51} &    &  \\
\rowcolor[rgb]{0.851, 0.851, 0.851} \citet{sv_deep2024} (2024) & \ding{51} & \ding{51} & \ding{51} & \ding{51} & & & & \ding{51} & \ding{51} & & \\
Ours &   \ding{51} & \ding{51} &  \ding{51}  &  \ding{51}  &  \ding{51}  & \ding{51} & \ding{51} & \ding{51} &  \ding{51}  &  \ding{51}  & \ding{51} \\
\bottomrule
\end{tabular}
\label{tab:survey}
}
\end{table*}

Table~\ref{tab:survey} compares our work with recent key related surveys to position our contributions within the existing literature. A detailed comparison of all related surveys is provided in SM Table~\ref{tab:survey_full}.
As the table illustrates, our survey is distinguished from prior studies in three key aspects:
\begin{itemize}
    \item \textbf{Broad Coverage of Emerging Architectures:} Existing surveys rarely cover the full spectrum of modern deep learning architectures, such as prototype networks, state space models, diffusion models, and LLMs (as detailed in the \textit{Novel Algorithms} column of Table~\ref{tab:survey}). Cross-domain unified TSAD models also remain largely unexplored. This review provides a detailed overview of these cutting-edge algorithms and their operational characteristics.
    \item \textbf{Principled Taxonomy of Paradigms:} To accommodate the continuous influx of new models, we formalize a structured taxonomy that classifies algorithms according to their core detection assumptions rather than superficial backbone implementations, with developmental trajectories detailed in Section~\ref{app:Related_Works} of the Supplementary Material.
    \item \textbf{Refined Multivariate Anomaly Taxonomy:} While prior surveys propose anomaly taxonomies \cite{sv_cookAnomalyDetectionIoT2020,sv_garciaReviewOutlierAnomaly2020,sv_shaukatReviewTimeSeriesAnomaly2021,sv_ae_CompositeF1Score}, they remain largely confined to univariate formulations. Because multivariate detection must explicitly account for inter-channel relationships \cite{ae_audibertUSADUnSupervisedAnomaly2020,cont_zhong2024simad,cont_zhong2024patchad}, we formalize channel-correlation anomalies to provide a unified framework for complex cross-channel interactions.
\end{itemize}
\subsection{Definitions}
This subsection establishes formal mathematical notations and problem definitions for time series anomaly detection and multi-domain unified modeling, providing the foundation for the taxonomies introduced in Secs.~\ref{sec:Refined_Anomaly_Taxonomy} and \ref{sec:Taxonomy_of_Paradigm}.

\subsubsection{Definition of Problems}
Time series data comprises a sequence of observations ordered chronologically. Formally, a time series is represented as \(\boldsymbol{t} = [\boldsymbol{t}_1, \boldsymbol{t}_2, \dots, \boldsymbol{t}_N]^\top \in \mathbb{R}^{N \times C}\), where \(\boldsymbol{t}_i \in \mathbb{R}^C\) denotes the observation vector at timestamp \(i\), and \(N\) signifies the total sequence length. Each observation encompasses \(C\) distinct channels or variables. The univariate sequence for the \(c\)-th channel is denoted as \(\boldsymbol{t}_{:,c} = [t_{1,c}, t_{2,c}, \dots, t_{N,c}]^\top \in \mathbb{R}^{N}\). A sequence with \(C > 1\) is termed a \textit{multivariate time series}, whereas \(C = 1\) defines a \textit{univariate time series}.

While time series analysis conventionally assumes temporal continuity and strong auto-correlation, empirical datasets frequently deviate from ideal regularities. In practical deployments, detectors must contend with irregularities such as non-uniform sampling intervals \cite{fu2025hadn} and missing values \cite{yang2024multivariate}, which require specialized pre-processing or robust architectural accommodations.

Each timestamp is associated with a ground-truth binary label \(y_i \in \{0, 1\}\), where \(y_i = 1\) denotes an anomaly and \(y_i = 0\) indicates nominal behavior, yielding a label sequence \(\boldsymbol{y} = [y_1, y_2, \dots, y_N]^\top \in \{0, 1\}^N\). Because anomalies are inherently rare in real-world systems, anomalous observations are heavily outnumbered by nominal points. The objective of a TSAD algorithm is to estimate an anomaly score sequence and predict the binary label sequence \(\boldsymbol{y}\) from the input observations \(\boldsymbol{t}\).

\textbf{Anomaly Length}: 
A \textit{point anomaly} manifests at index \(i\) when \(y_i = 1\), where all existing immediate adjacent timestamps of \(i\) are normal. For boundary positions at the start or end of the entire sequence, only the single existing neighbor needs to be normal to form a point anomaly. 
A \textit{segment anomaly} (or sequence anomaly) spans a contiguous interval from \(i\) to \(j\) where \(y_{i:j} = 1\) \cite{sv_deep2024}. This definition does not require the timestamps immediately before \(i\) or after \(j\) to be normal, allowing distinct anomalies to exist in close proximity. The anomaly length is defined as \(L = j - i + 1\). Accounting for anomaly length is essential, as differing durations often reflect distinct operational severity levels \cite{ETS-Aware}.
\subsubsection{Multi-Domain Learning}
In time series analysis, multi-domain learning involves extracting shared representations across diverse temporal domains simultaneously \cite{domain_Lu2023OutofDistribution,domain_shiOrthogonalityMattersInvariant2024}. Formally, a multi-domain temporal dataset is defined as
\(\gD = \{ \gT, \gY, \gS \}\),
where \(\gT\), \(\gY\), and \(\gS\) represent the collections of time series observations, corresponding ground-truth labels, and domain identifiers, respectively. Here, \(\gT = \{ \boldsymbol{T}^1, \boldsymbol{T}^2, \dots, \boldsymbol{T}^{N_{d}} \}\) comprises data from \(N_{d}\) distinct domains, with each \(\boldsymbol{T}^i \in \mathbb{R}^{N_i \times C_i}\) encapsulating the observations specific to domain \(i\).
Correspondingly, the label set is \(\gY = \{ \vy_1, \vy_2, \dots, \vy_{N_{d}} \}\), where each domain's label sequence is given by
\(
\vy_{i} = \{ \evy_1, \evy_2, \dots, \evy_{N_i} \} \in \{0, 1\}^{N_i}.
\)
\added{Distinct domains commonly present heterogeneous sequence lengths \(N_i\) and channel dimensions \(C_i\), such that \(N_i \ne N_j\) or \(C_i \ne C_j\) holds for arbitrary domain pairs \(i \ne j\). Models designed for multi-domain settings must dynamically accommodate this structural heterogeneity while leveraging invariant cross-domain temporal knowledge.}

\added{Unified TSAD models are designed to detect anomalies across heterogeneous temporal domains through a single shared network \cite{cont_shentuGeneralTimeSeries2024}. Formally, given a multi-domain dataset \(\gD\), a unified detector \(\mathcal{M}\) maps temporal inputs \(\gT\) to anomaly scores \(\gA\) and domain-aware latent representations \(\gZ\):
\[
\mathcal{M}(\gT; \Theta, \Phi) \rightarrow (\gA, \gZ),
\]
where \(\Theta\) denotes domain-shared parameters that encode universal temporal features, and \(\Phi = \{\phi_1, \phi_2, \dots, \phi_{N_d}\}\) represents optional domain-specific adapters. The inclusion of \(\Phi\) depends on whether target domains are known during training.}

\added{An effective unified architecture realizes dual capabilities:
1) \textit{Parameter Sharing}: The core backbone \(\mathcal{M}\) relies primarily on the unified parameter set \(\Theta\) across all domains in \(\gD\), minimizing parameter redundancy and training overhead \cite{llm_UNITS2024}.
2) \textit{Dual-Level Generalization}: The model achieves both \textit{intra-domain} robustness (detecting anomalies under temporal distribution shifts within known domains) and \textit{cross-domain} transferability (generalizing to unseen target domains with zero or few-shot adaptation).
By harmonizing heterogeneous inputs into a shared representation space while maintaining high parameter efficiency, unified architectures provide a scalable foundation for modern industrial monitoring \cite{llm_UNITS2024}.}
\section{Refined Anomaly Taxonomy}\label{sec:Refined_Anomaly_Taxonomy}
An anomaly in temporal data represents a significant deviation from the underlying generative distribution, manifesting either as an isolated point observation or as a contiguous subsequence departing from nominal behavior \cite{sv_laiRevisitingTimeSeries}.

Historically, literature in time series analysis \cite{sv_braeiAnomalyDetectionUnivariate2020,sv_liDeepLearningAnomaly2023b,sv_laiRevisitingTimeSeries,sv_zhangTimeSeriesAnomaly2021} has categorized anomalies into global, contextual, and structural forms. However, this classical taxonomy is predominantly tailored for univariate sequences and fails to articulate how anomalies manifest across interdependent variables. This section first reviews these three classical univariate anomaly types and then, in Sec.~\ref{sec:cc_anomaly}, formalizes the taxonomy of channel-correlation (CC) anomalies to capture the intricate inter-channel dynamics governing multivariate systems.
\begin{figure*}[ht]
    \centering
    \includegraphics[width=\linewidth]{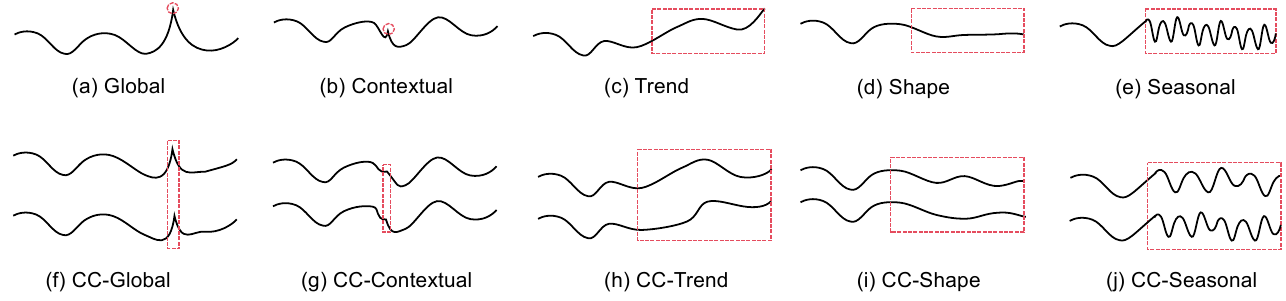}
    \caption{Illustration of different types of anomalies. The red box indicates the area of anomaly.}
    \label{fig:anomlies}
\end{figure*}
\subsection{Global Anomalies}
\added{\textbf{Global anomalies} represent extreme deviations within a time series, frequently manifesting as sudden spikes or abrupt level shifts that fall far outside the nominal distribution bounds \cite{sv_braeiAnomalyDetectionUnivariate2020,sv_deep2024}, as illustrated in Fig.~\ref{fig:anomlies}(a). Formally, a global anomaly at timestamp $i$ for channel $c$ is evaluated as:}
\begin{equation}
    \mathcal{F}_{ga} = |t_{i,c}-\hat{t}_{i,c}| > \delta_{ga},
\end{equation}
where $t_{i,c}$ denotes the observed value at timestamp $i$ for channel $c$, $\hat{t}_{i,c}$ is the expected or reconstructed baseline value, $\mathcal{F}_{ga}$ denotes the residual-based anomaly indicator, and $\delta_{ga}$ is the corresponding detection threshold. An observation is flagged as a global anomaly whenever its absolute deviation surpasses $\delta_{ga}$.

\subsection{Contextual Anomalies}
\added{\textbf{Contextual anomalies} manifest as observations whose values fall within the global range of the series but deviate significantly when evaluated against their localized temporal neighborhood \cite{v1_proof1,sv_deep2024,sv_braeiAnomalyDetectionUnivariate2020}, as shown in Fig.~\ref{fig:anomlies}(b). Contextual detection relies on surrounding temporal structure and is formally expressed as:}
\begin{equation}
    \mathcal{F}_{ca} = |\sigma(\boldsymbol{t}_{(i-m+1):i,\, c})-\sigma(\hat{\boldsymbol{t}}_{(i-m+1):i,\, c}) | > \delta_{ca},
\end{equation}
\added{where $\boldsymbol{t}_{(i-m+1):i,\, c}$ denotes the sequence of channel $c$ over a temporal window of length $m$ ending at index $i$, and the function $\sigma(\cdot)$ extracts a localized contextual feature, such as local variance or neighborhood density.}

\subsection{Structural Anomalies}
Unlike point-level irregularities, structural anomalies disrupt the underlying morphological or generative patterns of a time series across an extended temporal window, as illustrated in Figs.~\ref{fig:anomlies}(c--e). These anomalies generally encompass three primary forms:
\begin{itemize}
    \item \textbf{Trend anomalies} represent unexpected regime shifts or persistent drift in the low-frequency trajectory of a sequence \cite{sv_braeiAnomalyDetectionUnivariate2020} (Fig.~\ref{fig:anomlies}(c));
    \item \textbf{Shape anomalies} occur when local temporal waveforms violate expected morphological patterns, even while point values remain within nominal bounds (Fig.~\ref{fig:anomlies}(d));
    \item \textbf{Seasonal anomalies} involve deviations from regular periodic cycles, where timing or phase relationships are distorted despite normal cyclical amplitudes (Fig.~\ref{fig:anomlies}(e)).
\end{itemize}
These structural variations can be unified under a generalized evaluation formula:
\begin{equation}
    \mathcal{F}_{sa} = \operatorname{StrucFunc}_{\text{type}}(\boldsymbol{t}_{(i-m+1):i,\, c},\ \hat{\boldsymbol{t}}_{(i-m+1):i,\, c}) > \delta_{sa},
\end{equation}
\added{where $\operatorname{StrucFunc}_{\text{type}}$ evaluates deviations across decomposed representations, such as trend trajectories, seasonal profiles, or shapelet distances, against detection threshold $\delta_{sa}$.}

\subsection{Channel-Correlation Anomalies}\label{sec:cc_anomaly}
While previous literature has established diverse anomaly taxonomies \cite{sv_braeiAnomalyDetectionUnivariate2020,sv_liDeepLearningAnomaly2023b,sv_laiRevisitingTimeSeries,sv_zhangTimeSeriesAnomaly2021}, existing studies rarely formalize the boundary between univariate and multivariate anomalies, nor do they quantify how cross-channel dependencies govern the composite anomaly score. Many existing methods \cite{sv_laiRevisitingTimeSeries,trs_yangDCdetectorDualAttention2023,ae_audibertUSADUnSupervisedAnomaly2020} adopt the channel-independence assumption, detailed in Section~\ref{app:general_technologies} of the Supplementary Material, which treats each variable as an isolated series. However, this assumption remains valid only when inter-channel correlations are demonstrably negligible \cite{sv_deep2024}.

Building upon the concept of correlation anomalies introduced by \citet{v1_proof1}, we formalize a unified taxonomy: Channel-correlation (CC) anomalies originate from synchronous or coupled abnormal behaviors across multiple channels, manifesting as disruptions in the joint distribution rather than extreme marginal values in individual variables. As demonstrated by \citet{roy2025beyond}, subtle multivariate anomalies often elude single-channel inspection because individual variables appear statistically nominal when viewed in isolation. Such events become identifiable only through cross-channel joint modeling. Consequently, contemporary research increasingly deploys joint models that explicitly capture inter-variable dependencies.

Figs.~\ref{fig:anomlies}(f--j) illustrate five distinct classes of channel-correlation anomalies: CC-Global, CC-Contextual, CC-Trend, CC-Shape, and CC-Seasonal. In Fig.~\ref{fig:anomlies}(g), although the peaks in the two channels individually surpass their contextual thresholds, it is their synchronous co-occurrence that constitutes the multivariate anomaly. In Fig.~\ref{fig:anomlies}(i), the individual shapes of both channels adhere largely to historical profiles, leading marginal detectors to misclassify the interval as nominal. Under the CC-Shape formulation, however, the coupled shape distortion across both channels collectively flags the anomaly.

We formalize the anomaly score for a single channel \(c\) as \(\mathcal{F}_{\text{type}}^c\). Channel-correlation anomalies can be defined as follows:
\begin{equation}
\begin{aligned}
    &\mathcal{F}_{\text{cca}} = \sum_{c=1}^{C} \mathcal{F}_{\text{type}}^c > \sigma_{\text{cca}}\\
    &\text{s.t. } \forall \text{type} \in \{\text{ga}, \text{ca}, \text{sa}\}, \forall c, \quad \mathcal{F}_{\text{type}}^c > \sigma_{\text{type}}.
    \label{eq:cca1}
\end{aligned}
\end{equation}
This formulation flags a channel-correlation anomaly when every channel-specific score \(\mathcal{F}_{\text{type}}^{c}\) exceeds its individual threshold \(\sigma_{\text{type}}\) and their aggregate sum simultaneously surpasses the joint threshold \(\sigma_{\text{cca}}\). This concurrent violation captures anomalous dynamics unique to coupled multivariate sequences.

In practical deployments, isolating whether a multivariate event is driven by a specific underlying anomaly sub-type per channel is often intractable. A generalized and widely adopted formulation evaluates each channel's weighted contribution to the joint anomaly score:
\begin{equation}
\begin{aligned}
    \mathcal{F}_{\text{cca}} &= \sum_{c=1}^{C} \alpha_c\,\mathcal{F}^{c}\;>\;\sigma_{\text{cca}},\\
    &\text{with}\;\forall c,\; \mathcal{F}^{c}>\sigma_{c},
\end{aligned}
\label{eq:cca2}
\end{equation}
where \(\alpha=[\alpha_1,\dots,\alpha_C]\) denotes predefined channel importance weights, and \(\mathcal{F}^{c}\) represents the aggregate anomaly score for channel \(c\). The weights $\alpha_c$ enable uninformative or decoupled channels to be masked by setting $\alpha_c = 0$ when an anomaly is confined to a specific functional subsystem.

While (\ref{eq:cca1}) and (\ref{eq:cca2}) define scoring rules to detect cross-channel synchrony, they serve as empirical identification criteria rather than explicit generative models. In operational environments, multivariate anomalies emerge from diverse inter-channel dynamics; our formulation prioritizes observable synchrony to ensure robust practical detection.

In critical cyber-physical systems and autonomous platforms, multivariate time series are additionally governed by physical and kinematic constraints \cite{meng2025mars}. Whereas classical model-based detectors naturally incorporate kinematic equations to identify physically inadmissible trajectories, data-driven model-free methods require explicit inductive biases or structured embeddings. Recent architectures address this requirement through physics-informed deep learning, such as Pi-Transformer \cite{maleki2025pitransformer}, which embeds conservation laws directly into the attention mechanism to ensure that detected deviations reflect genuine physical faults rather than unconstrained modeling artifacts.
\section{Taxonomy of Paradigms}\label{sec:Taxonomy_of_Paradigm}
In this section, we categorize existing TSAD algorithms according to their core operational paradigms, outlining their fundamental methodologies and formal structures.

\begin{figure*}[ht]
    \centering
    \includegraphics[width=1\linewidth]{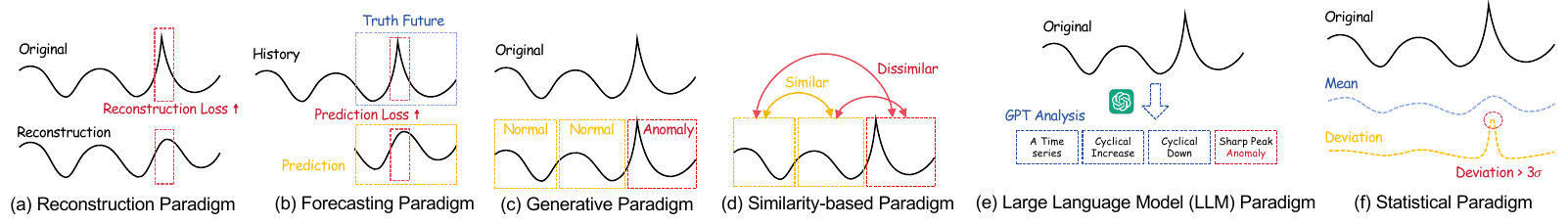}
    \caption{Diagram of different paradigms of TSAD.}
    \label{fig:paradigms}
\end{figure*}

\added{To ensure unambiguous paradigm assignment, the taxonomy is grounded in a three-tier classification framework. Detailed decision criteria, a boundary disambiguation table, and illustrative case analyses are provided in SM~\ref{sec:paradigm_criteria}.}
\begin{enumerate}
    \item \added{\textbf{LLM Priority Rule.} The emergence of foundational models represents a paradigm shift from learning task-specific distributions from scratch to leveraging pre-trained cross-domain knowledge for zero-shot generalization. Accordingly, methods whose detection capability fundamentally relies on such pre-trained representations are assigned to the LLM-based Paradigm, irrespective of the mathematical form of their anomaly score.}
    \item \added{\textbf{Core Detection Assumption.} All other methods are classified by their core detection assumption, which defines the fundamental criterion for identifying anomalies across Sec.~\ref{sec:rec_para} through Sec.~\ref{sec:stat_pard}.}
    \item \added{\textbf{Dominant Mechanism Rule.} When a method employs multiple scoring signals, it is assigned to the paradigm of its primary anomaly score; hybrid cases are noted in SM~\ref{sec:paradigm_criteria}.}
\end{enumerate}
\added{Fig.~\ref{fig:paradigms} illustrates the conceptual detection principles of each paradigm. Complementarily, Table~\ref{tab:algo_cite_table} compiles representative literature across paradigms, neural backbones, and unified models, with detailed algorithmic treatments in SM~\ref{sec:Algorithm_Innovations}. The alignment between architectural backbones and detection paradigms is further analyzed in Sec.~\ref{sec:method_choice} and illustrated in Fig.~\ref{fig:alg_ch1}.}

\newlength{\mycolwidth}
\setlength{\mycolwidth}{3cm}
\begin{table}[htbp]
  \centering
  \caption{Summary of algorithmic innovations. Detailed descriptions and references are available in SM~\ref{sec:Algorithm_Innovations}.}
  \resizebox{\linewidth}{!}{
\begin{tabular}{p{\mycolwidth}|p{3.2cm}|p{8.5cm}}
\toprule
\textbf{Dimension} & \textbf{Category} & \textbf{Representative Algorithms} \\
\midrule
\multirow{6}[8]{=}{Detection Paradigms (SM~\ref{sec:Taxonomy_of_Paradigms_Methods})} 
& Reconstruction & EncDec-AD, MSCRED, UAE, G-LSTM-AE, USAD, RE-ADTS, NPSR \\
\cmidrule{2-3} & Forecasting & ARIMA, FEDformer, Autoformer, PatchTST, DLinear \\
\cmidrule{2-3} & Generative & LSTM-VAE, InterFusion, MUTANT, FCVAE, FGANomaly, TadGAN, AnoFormer, BeatGAN, UAC-AD, D3R, ImDiffusion, DDMT, TimeADDM, DDTAD \\
\cmidrule{2-3} & Similarity-based & KNN, DTW, PSOM, LOF, HBOS, ECOD, OC-TCN, COCA, COUTA, THOC, PatchAD, CARLA, SimAD, TimeAutoAD, USD, BroadCAE, MEMTO, AMSL, AdaMemBLS, OCSVM, EB-OCSVM, Isolation Forest \\
\cmidrule{2-3} & Statistical & Z-Scores, 3-Sigma, IQR \\
\cmidrule{2-3} & LLM-based & TS-Bert, GPT4TS, aLLM4TS, Timer, AnomalyLLM, LLMAD, SIGLLM, ParInfoGPT, Holmes-VAD, MERIT \\
\midrule
\multirow{6}[16]{=}{Representation Architectures (SM~\ref{sec:Deep_Learning_Architecture})} 
& RNN Backbone & OmniAnomaly, CAE-AD, MSDMM, LSGA \\
\cmidrule{2-3} & GNN Backbone & STADN, DGINet, DPGLAD, SGTrans \\
\cmidrule{2-3} & TCN Backbone & TCN-AE, WaRTEm-AD, TFAD \\
\cmidrule{2-3} & State Space Model & S-Mamba, DMamba, C-Mamba, Mambaformer, SGFM, TSAD-C \\
\cmidrule{2-3} & Transformer Backbone & Anomaly Transformer, TransAnomaly, CauseFormer, TiSAT, TranAD, AnomalyBERT, DCdetector, D3TN \\
\cmidrule{2-3} & Others & Graph-Augmented Normalizing Flows, PeFAD \\
\midrule
{Unified TSAD Models (SM~\ref{sec:Unified_TSAD})} & Cross-Domain Models & DADA, DACAD, ContexTDA, FS-ADAPT, VANDA, UNITS \\
\bottomrule
\end{tabular}}
  \label{tab:algo_cite_table}%
\end{table}

\subsection{Reconstruction Paradigm}\label{sec:rec_para}
As illustrated in Fig.~\ref{fig:paradigms}(a), the reconstruction paradigm exploits the intrinsic structural regularities of time series, focusing on temporal continuity and inter-channel correlations. The core premise posits that normal temporal patterns can be faithfully reconstructed from compressed latent representations, whereas substantial reconstruction errors signal potential abnormalities \cite{sv_deep2024,trs_yangDCdetectorDualAttention2023,cont_zhong2024simad}. 

Formally, for a given time sequence \(\boldsymbol{t}\), this paradigm employs an autoencoder framework defined as \(f_{\text{rec}}=\{f_{\text{enc}}, f_{\text{dec}} \}\). The encoder \(f_{\text{enc}} : \mathbb{R}^{m \times C} \rightarrow \mathbb{R}^{d}\) maps an input time window \(\boldsymbol{t}_{(i-m+1):i}\) of length \(m\) into a \(d\)-dimensional latent feature \(\boldsymbol{z}\):
\begin{equation}
   \boldsymbol{z} = f_{\text{enc}}(\boldsymbol{t}_{(i-m+1):i}). 
\end{equation}
Subsequently, the decoder \(f_{\text{dec}} : \mathbb{R}^{d} \rightarrow \mathbb{R}^{m \times C}\) reconstructs the original temporal window from this latent representation:
\begin{equation}
   \hat{\boldsymbol{t}}_{(i-m+1):i} = f_{\text{dec}}(\boldsymbol{z}). 
\end{equation}

The reconstruction error, typically quantified via Mean Squared Error (MSE) \cite{ae_laiNominalityScoreConditioned2023a,ae_amarbayasgalanUnsupervisedAnomalyDetection2020}, serves as the anomaly score:
\begin{equation}
   \mathcal{F}_{i}^{\text{rec}} = \frac{1}{m} \sum_{k=i-m+1}^{i} \|\boldsymbol{t}_k - \hat{\boldsymbol{t}}_k\|_2^2. 
\end{equation}
Anomalies are flagged when \(\mathcal{F}_i^{\text{rec}}\) exceeds a predefined threshold \(\delta\).

\subsection{Forecasting Paradigm}\label{sec:pred_pard}
As illustrated in Fig.~\ref{fig:paradigms}(b), the forecasting paradigm prioritizes predicting forthcoming temporal values rather than reconstructing historical observations \cite{tcn_huangEfficientTimeSeries2022,llm_UNITS2024}. In the literature, this approach is alternatively termed \textit{predictability modeling} or \textit{prediction-based detection} \cite{ae_gargEvaluationAnomalyDetection2022,sv_deep2024,sv_zhangTimeSeriesAnomaly2021}. This methodology posits that predictive models capture normal evolutionary dynamics, such that unexpected anomalies manifest as prominent discrepancies between forecasted trajectories and actual observations.

Formally, a forecasting model \( f_{\text{pred}} : \mathbb{R}^{m \times C} \rightarrow \mathbb{R}^{p \times C} \) takes a historical observation window \(\boldsymbol{t}_{(i-m+1):i}\) of length \(m\) to predict the subsequent \(p\) time steps:
\begin{equation}
\hat{\boldsymbol{t}}_{(i+1):(i+p)} = f_{\text{pred}}(\boldsymbol{t}_{(i-m+1):i}).
\end{equation}
The prediction error, typically quantified using MSE, acts as the anomaly score for the forecasted horizon:
\begin{equation}
    \mathcal{F}_{i}^{\text{pred}} = \frac{1}{p} \sum_{k=i+1}^{i+p} \|\boldsymbol{t}_k - \hat{\boldsymbol{t}}_k\|_2^2.
\end{equation}
An anomaly is triggered when this discrepancy exceeds a designated threshold.

Recent empirical evaluations \cite{timemixer_PP_2025,cont_zhong2024patchad,trs_yangDCdetectorDualAttention2023,TSB-AD} demonstrate that architectures originally designed for pure forecasting can effectively perform TSAD out-of-the-box, provided their intermediate layers extract stable and consistent temporal representations.

\subsection{Generative Paradigm}\label{sec:gen_pard}
As depicted in Fig.~\ref{fig:paradigms}(c), the generative paradigm models the underlying probabilistic generation process of normal sequences. It fits a generative model to the training set and flags observations exhibiting unusually low likelihoods or substantial deviations from synthetic counterparts \cite{gan_duGANBasedAnomalyDetection2021,gan_geigerTadGANTimeSeries2020,gan_liuTimeSeriesAnomaly2023,gan_zengMultivariateTimeSeries2023a}. While often incorporating reconstruction-like mechanisms during processing, its core detection principle fundamentally relies on stochastic likelihood or density estimation under a learned data distribution, rather than deterministic point-wise feature mapping. In addition, generative approaches uniquely possess the capability to explicitly sample and synthesize new data from the learned continuous distribution, distinguishing them from strictly discriminative reconstruction models.

Formally, a generative model \( G : \mathbb{R}^{d} \rightarrow \mathbb{R}^{C} \) (or across a full temporal window \(\mathbb{R}^{m \times C}\)) maps a latent variable \(\boldsymbol{z}\) to the observation space to produce synthetic data \(\hat{\boldsymbol{t}}\). During training, the parameters \(\theta\) are optimized to approximate the true data distribution, frequently through maximum likelihood estimation or equivalent objective functions:
\begin{equation}
    \mathcal{L}(\theta) = \mathbb{E}_{\boldsymbol{t} \sim p_{\text{data}}, \boldsymbol{z} \sim p(\boldsymbol{z})} \left[ \mathcal{D}\big(\boldsymbol{t}, G(\boldsymbol{z}; \theta)\big) \right],
\end{equation}
where \(p_{\text{data}}\) represents the empirical data distribution and \(p(\boldsymbol{z})\) is the latent prior. 

During inference, for an observed data point \(\boldsymbol{t}_i\), a corresponding latent representation \(\boldsymbol{z}_i\) is inferred or sampled, and the synthetic observation \(\hat{\boldsymbol{t}}_i = G(\boldsymbol{z}_i; \theta)\) is generated. The anomaly score is determined by a discrepancy metric \(\mathcal{D}\):
\begin{equation}
\mathcal{F}_{i}^{\text{gen}} = \mathcal{D}(\boldsymbol{t}_i, \hat{\boldsymbol{t}}_i). 
\end{equation}
Common choices for \(\mathcal{D}\) include MSE or Kullback-Leibler (KL) divergence \cite{sv_gan}. Alternatively, adversarial frameworks such as Generative Adversarial Networks (GANs) may employ a dedicated discriminator network to directly evaluate sample authenticity \cite{GAN}.

\added{This paradigm is distinguished from the Reconstruction Paradigm by its probabilistic training objective, optimizing stochastic objectives such as the evidence lower bound (ELBO), adversarial losses, or score-matching criteria rather than deterministic point-wise residuals alone. It is further distinguished from the Statistical Paradigm in that the data distribution is implicitly captured through neural network parameterization rather than specified in closed analytical form, as detailed in SM~\ref{sec:paradigm_criteria}.}

\subsection{Similarity-based Paradigm}\label{sec:sim_pard}
As illustrated in Fig.~\ref{fig:paradigms}(d), the similarity-based paradigm identifies anomalies by explicitly computing geometric or statistical proximity measures across data samples.

Formally, given a time series \(\boldsymbol{t}\), individual time points or local windows are treated as distinct samples. A generalized similarity function \(\operatorname{sim}(\cdot, \cdot)\) is defined to quantify pairwise proximity. Depending on the underlying mathematical formulation, these comparative strategies fall into four primary operational mechanisms, all positing that anomalies exhibit low similarity to the predominant data distribution:
\begin{enumerate}
\item \textbf{Distance-based Evaluation}: \textit{Anomalies are isolated points situated at substantial distances from the majority cluster.} Employing metrics such as Euclidean distance or Dynamic Time Warping (DTW), models flag points that violate designated proximity thresholds.
\item \textbf{Density-based Evaluation}: \textit{Normal observations populate high-density regions, whereas anomalies reside in sparse neighborhoods.} Algorithms such as Local Outlier Factor (LOF) compute local density estimates, assigning elevated anomaly scores to points lacking dense neighbor clusters.
\item \textbf{Classification-based Evaluation}: \textit{Anomaly detection operates as an unsupervised boundary-learning task.} Methods such as One-Class Support Vector Machines (OCSVM) construct high-dimensional hyperspheres encapsulating normal data, classifying points lying outside this boundary as anomalies.
\item \textbf{Tree-structured Implementation}: \textit{Anomalies are inherently susceptible to rapid isolation due to their low-density nature.} Architectures such as Isolation Forest recursively partition the feature space; anomalous points require fewer splits, yielding shorter tree path lengths that directly reflect their low similarity to normal dense clusters.
\end{enumerate}
\added{Density-based methods, such as LOF and HBOS, rely on non-parametric neighborhood geometry, distinguishing them from the Statistical Paradigm which fits closed-form distributional parameters. Contrastive and prototype-based approaches, though employing neural feature extractors, are assigned here because their anomaly score derives from geometric distance in embedding space rather than from the backbone architecture.}

\subsection{LLM-based Paradigm}\label{sec:llm_pard}
Distinct from classical statistical or connectionist models that learn task-specific error distributions from scratch, the LLM-based paradigm (Fig.~\ref{fig:paradigms}(e)) introduces a fundamentally distinct detection principle: knowledge-driven semantic inference. By leveraging pre-trained cross-domain historical contexts and zero-shot reasoning capabilities, this paradigm transitions the detection focus from computing traditional mathematical error thresholds to formulating semantic, context-aware judgments \cite{llm_aLLM4TS2024,llm_AnomalyLLM2024,llm_ContextTDA_2023,llm_GPT4TS2023,llm_LLMAD2024,llm_DACAD2024,llm_ParInfoGPT2024}. Capitalizing on the autoregressive nature of foundational models, adaptations for TSAD generally align representations and capture temporal context via three primary strategies:

\textbf{1) Injecting External Knowledge}: For zero-shot TSAD tasks, external domain knowledge is directly injected into frozen LLMs via advanced prompt engineering or Retrieval-Augmented Generation (RAG) \cite{llm_TS-Bert2021,llm_aLLM4TS2024}. Because LLMs map \(\mathsf{text} \rightarrow \mathsf{text}\), numerical time series undergo a serialization process \(\mathsf{Inject}: \mathsf{time\ series} \rightarrow \mathsf{text}\) that transforms raw arrays into descriptive text. For example, a numerical sequence \([1, 3, 5, 10]\) is translated into the sentence: \textit{``The series increases from 1 to 10 via intermediate values 3 and 5.''} This serialized representation is then paired with a task-oriented system prompt instructing the model to analyze sequence patterns, identify anomalous intervals, and generate diagnostic explanations. Guided by these prompts, the LLM leverages its natural language generation capabilities to pinpoint and explain anomalies through semantic alignment \cite{llm_aLLM4TS2024}.

\textbf{2) Parameter-Efficient Fine-Tuning Strategies}: To enable few-shot adaptation, this strategy freezes core LLM layers, such as multi-head self-attention and feedforward networks containing extensive pre-trained knowledge, fine-tuning only positional embeddings and layer normalization \cite{llm_GPT4TS2023}. This adapts the input embedding layer to project numerical sequences into tokenized representations, effectively balancing broad pre-trained cross-domain representations with dataset-specific alignment.

\textbf{3) Multimodal and Agentic Extensions}: Moving beyond strictly sequential or tokenized adaptations, recent innovations harness multimodal large language models (MLLMs) and agentic workflows to enrich the representation space of TSAD. For instance, vision-language models can interpret time series data converted into visual charts or plots, thereby exploiting their mature, pre-trained pattern-recognition capabilities \cite{he2026harnessing}. In complex industrial or surveillance contexts, MLLMs jointly align and analyze heterogeneous streams, including numerical sensors, textual logs, and visual imagery \cite{mokhtar2025detect,zhang2024holmes}. Furthermore, conceptualizing LLMs as reasoning engines enables multi-agent collaborative frameworks, where specialized agents interact iteratively to refine unsupervised representations, handle domain-specific abstractions, and execute complex TSAD tasks \cite{zhou_merit_2025}.

\added{Under this classification, methods such as GPT4TS~\cite{llm_GPT4TS2023} and Timer~\cite{llm_TIMER2024}, whose anomaly scores take the form of prediction errors but whose detection capability derives from pre-trained cross-domain representations, are assigned here rather than to the Forecasting Paradigm.}

\subsection{Statistical Paradigm}\label{sec:stat_pard}
As illustrated in Fig.~\ref{fig:paradigms}(f), the statistical paradigm identifies anomalies by fitting explicit probability distributions or descriptive summary statistics directly to observed temporal sequences. \added{A defining characteristic of this paradigm is that the distributional model takes a closed-form analytical expression whose parameters are estimated without neural optimization, distinguishing it from the Generative Paradigm where the distribution is implicitly encoded in network weights.} This typically involves decoupling the periodicity and trend information in time series sequences. Traditional parametric techniques, such as box plots and Z-scores, evaluate anomalies by calculating summary statistics such as the mean and standard deviation. Moving beyond simple distance-based thresholds, log-probability serves as a more expressive metric within this paradigm. In parametric density-based models, such as multivariate Gaussian distributions or Gaussian mixture models, the detector estimates the joint probability distribution of the sequence, flagging observations that exhibit unusually low log-density as anomalies.
\section{Discussions}\label{sec:Discussions}
This section examines evaluation methodologies, architectural selection criteria, and foundational inductive biases in TSAD, identifying critical bottlenecks and outlining actionable research directions.

\subsection{Discussions on Evaluations}\label{sec:dis_eval}
Evaluating TSAD algorithms requires addressing two interdependent mechanisms: threshold determination and performance metric selection. We subsequently present an experimental analysis assessing the discriminability and stability of diverse evaluation metrics.

\subsubsection{Anomaly Threshold}
Thresholds demarcate nominal observations from anomalous deviations. Conventional approaches predominantly rely on static thresholds established as invariant classification boundaries. While computationally straightforward, static cutoffs degrade rapidly under temporal non-stationarity. In high-frequency operational or financial monitoring, for example, fixed thresholds fail to accommodate structural behavioral shifts triggered by external regime changes. Although input normalization standardizes the marginal distribution \( P(X) \), it cannot ensure invariant scoring behavior. Because anomaly scores depend on the learned conditional mapping \( P(Y|X) \) \cite{long2013transfer}, distributional shifts in this conditional mapping induced by evolving operational contexts alter score distributions, rendering static thresholds brittle.

This vulnerability underscores the necessity of dynamic thresholding mechanisms capable of continuous adaptation. However, dynamic calibration remains substantially underexplored in TSAD literature. Future investigations should prioritize end-to-end learnable thresholds that infer scenario-specific decision boundaries directly from operational context, eliminating arbitrary heuristic cutoffs.

\subsubsection{Flawed Metrics}\label{sec:flaw_metric}
The reliability of comparative evaluations hinges on the choice of scoring metrics. While earlier benchmarks relied primarily on the point-adjustment F1 (F1-PA) score in conjunction with AUC-ROC and AUC-PR, recent theoretical and empirical analyses \cite{VUS-ROC,affiliation,2025arXiv251116145Z} have exposed the severe inflation bias inherent to F1-PA. Specifically, under point adjustment, correctly detecting a single timestamp within an anomalous segment automatically labels the entire segment as successfully identified. This mechanism artificially inflates precision and recall, allowing even random score generators to achieve deceptively competitive performance on segment anomalies. 

Although the community has explored range-aware alternatives, a universally accepted evaluation standard remains elusive. Algorithms frequently optimize for specific metric dimensions without demonstrating consistent superiority across all evaluation axes, precluding rigorous cross-study comparability. 

To quantitatively demonstrate the inflation bias of F1-PA against alternative protocols, we present a synthetic case study in SM~\ref{app:metric_case_analysis}. We strongly advise against relying on F1-PA in empirical benchmarks, as range-wise protocols provide substantially higher evaluation fidelity \cite{sv_sarfraz2024positionquovadisunsupervised,zhong2025cce}. Crucially, our experimental findings detailed in Sec.~\ref{sec:analysis_of_metric} confirm that no individual metric encapsulates performance across heterogeneous anomaly densities and sequence horizons \cite{cont_zhong2024simad,tsbuad,TSB-AD,sv_wuCurrentTimeSeries2022}. Robust benchmarking necessitates multi-metric evaluations accompanied by open-source implementations and pre-trained checkpoints to guarantee scientific reproducibility.

\subsubsection{Analysis of Metrics}\label{sec:analysis_of_metric}
\begin{parcolor}[\mycolor]
In this subsection, we experimentally evaluate the benchmark metrics summarized in Table~\ref{tab:comparison_metric}, focusing on discriminative power, computational latency, and distributional sensitivity across synthetic and real-world benchmarks.

Detailed experimental configurations and sensitivity evaluations across varied scoring backbones, sequence lengths, and segment durations are provided in SM~\ref{app:experimental_analysis_metrics}, corresponding to Figs.~\ref{fig:vs_model}--\ref{fig:vs_seg_len} and Table~\ref{tab:metric_ana_config}. These empirical analyses establish four primary findings:

\begin{itemize}
    \item \textbf{Robustness and Variance:} Point-wise F1 and F1-PA exhibit high vulnerability to score perturbations. F1-PA exhibits substantial score variance and unstable baseline bias, rendering it unsuited for rigorous performance differentiation.
    \item \textbf{Metric Discriminability:} Range-based metrics display markedly disparate discriminative capacities. Reduced-F1, R-based F1, TaPR, eTaPR, PATE, and AUC-PR frequently fail to separate detectors of distinct competency. Furthermore, R-based F1 and TaPR are structurally bounded below their theoretical maximums even under perfect detection.
    \item \textbf{Computational Latency:} Evaluation overhead scales steeply with sequence length and segment density. Because PATE incurs quadratic computational complexity, its deployment becomes computationally prohibitive on datasets exceeding 100,000 timestamps or containing over 50 anomalous segments.
    \item \textbf{Sensitivity to Segment Length:} Segment duration directly influences metric bias. For brief anomalies with minimal duration variance, metrics such as F1, PA\%K, Reduced-F1, TaPR, and AUC-PR present low distortion. Conversely, affiliation-based metrics (NAff-F1, UAff-F1) and volume-under-surface measures (VUS-ROC, VUS-PR) provide superior evaluation fidelity on extended anomalous intervals and high-variance regimes.
\end{itemize}

Actionable protocols for task-specific metric selection are consolidated in SM~\ref{app:practical_rec_4_metric_sel}.
\end{parcolor}

\subsection{Discussions on Algorithm Architectures}\label{sec:method_choice}
\begin{figure}[htbp]
    \centering
    \includegraphics[width=0.95\linewidth]{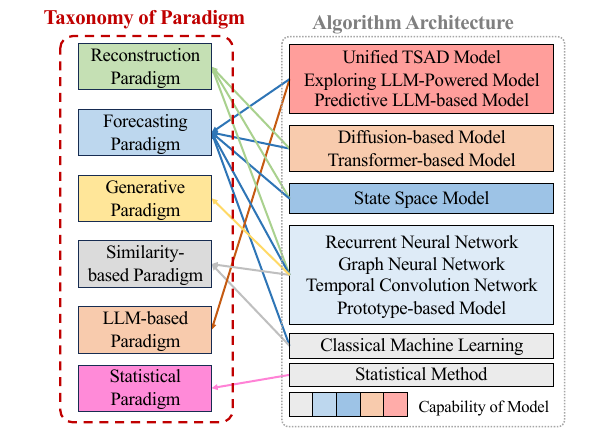}
    \caption{The mapping relationship between algorithm architectures (categorized by model capability) and the taxonomy of paradigms.}
    \label{fig:alg_ch1}
\end{figure}
Fig.~\ref{fig:alg_ch1} illustrates the bipartite mapping between detection paradigms and representation architectures. The \textit{Taxonomy of Paradigms} (left) formalizes mathematical assumptions and anomaly scoring principles, whereas the \textit{Algorithm Architecture} (right) organizes backbones along a spectrum of representational capacity, spanning closed-form statistical estimators to high-capacity foundational models capable of zero-shot reasoning. Color gradients in Fig.~\ref{fig:alg_ch1} signify this capacity progression, with directional edges marking predominant architectural implementations across paradigms. Detailed algorithmic treatments appear in SM~\ref{sec:Algorithm_Innovations}.

To delineate architectural trade-offs, Table~\ref{tab:alg_ch2} summarizes the representative strengths and limitations of each model class. Table~\ref{tab:alg_ch3} further translates these trade-offs into practical selection guidelines across operational constraints. Following the color coding of Fig.~\ref{fig:alg_ch1}, neutral gray denotes classical machine learning and statistical methods, while pink highlights high-capacity unified and LLM-based architectures. For each decision criterion, candidate architectures are categorized into recommended, secondary, and unsuitable tiers.

To guide practitioners in selecting appropriate detection backbones, we outline six structured recommendations:
\begin{enumerate}
    \item \textbf{Multimodal and Zero-Shot Generalization:} Prioritize unified and LLM-based architectures when tasks demand cross-domain knowledge transfer, multimodal contextual reasoning, or zero-shot inference across unseen domains.
    \item \textbf{Complex Temporal and Channel Dynamics:} Prioritize specialized deep representation architectures, such as Transformer or recurrent backbones \cite{ae_malhotraLSTMbasedEncoderDecoderMultisensor2016}, over classical machine learning models when capturing intricate long-range temporal dependencies and inter-variable interactions is essential.
    \item \textbf{Mathematical Transparency and Interpretability:} Deploy rigorous statistical models and analytical score indicators where decision transparency is non-negotiable. While ensemble methods provide post-hoc feature importance, statistical formulations offer direct, closed-form interpretability.
    \item \textbf{Resource-Constrained and Real-Time Streaming:} Favor lightweight statistical techniques and classical machine learning algorithms under stringent computational or latency budgets. Small-scale recurrent or temporal convolutional networks serve as viable deep learning alternatives balancing accuracy and throughput.
    \item \textbf{Peak Empirical Accuracy:} Adopt Transformer-based architectures when raw detection precision is the overriding benchmark criterion.
    \item \textbf{Baseline Selection Considerations:} Exercise caution when deploying diffusion, state space, or foundational models as baseline comparators, as their substantial training overhead and evolving design spaces make them less suited for minimal reference benchmarks.
\end{enumerate}

\begin{table*}
    \centering
    \caption{Comparative analysis of strengths and limitations across representation architectures.}
    \resizebox{0.93\linewidth}{!}{
\begin{tabular}{|p{14em}<{\centering}|p{23em}|p{23em}|}
    \toprule
    \multicolumn{1}{|c|}{\textbf{Architecture}} & \multicolumn{1}{c|}{\textbf{Advantages}} & \multicolumn{1}{c|}{\textbf{Disadvantages}} \\
    \midrule
    \rowcolor[rgb]{ 1,  .675,  .663} { \ \newline{}\newline{}Unified TSAD Model\newline{}Exploring LLM-Powered Model\newline{}Predictive LLM-based Model} & \ding{172}High representational capacity providing a versatile framework for cross-domain and multi-task learning.\newline{}\ding{173}Seamless integration with heterogeneous modalities (such as vision and text) via multimodal LLMs for context-rich reasoning.\newline{}\ding{174}Facilitates collaborative multi-agent architectures that leverage generative capabilities for automated diagnostic reasoning. & \ding{172}Substantial parameter footprint, demanding memory overhead, and elevated inference latency.\newline{}\ding{173}Evolving research domain without consistent empirical margins over specialized architectures.\newline{}\ding{174}Heavy reliance on textual tokenization or 2D image rendering, causing token bloat in high-frequency numerical time series. \\
    \midrule
     \rowcolor[rgb]{ .976,  .796,  .667} \ \newline{}\newline{}\newline{}Diffusion-based Model
     & \ding{172}Exceptional generative capacity for accurately capturing complex high-dimensional distributions.\newline{}\ding{173}High modeling flexibility with minimal parametric assumptions on data distributions. & \ding{172}Elevated computational overhead from iterative reverse denoising steps during both training and inference, limiting real-time utility.\newline{}\ding{173}High data dependency requiring substantial training volumes to faithfully learn normal temporal baselines. \\
     \midrule
     \rowcolor[rgb]{ .976,  .796,  .667} \ \newline{}Transformer-based Model
     & \ding{172}Highly effective at modeling non-local associations and pairwise variable relationships across extended sequences.\newline{}\ding{173}Fully parallelized training over entire temporal sequences, accelerating model convergence. 
     & \ding{172}Quadratic computational and memory complexity with respect to sequence length, creating scalability bottlenecks. \\
    \midrule
    \rowcolor[rgb]{ .616,  .765,  .902} \ \newline{}\newline{}\newline{} {State Space Model} & \ding{172}Linear computational complexity with sequence length, enabling efficient long-horizon temporal processing.\newline{}\ding{173}Favorable efficiency-performance trade-off compared to dense self-attention under equivalent parameter scales. & \ding{172}Relatively nascent architectural ecosystem with limited off-the-shelf tooling and pre-trained baselines.\newline{}\ding{173}Empirical advantages over long horizons require careful recurrent state parameterization and domain tuning. \\
    \midrule
    \rowcolor[rgb]{ .871,  .922,  .969} \ \newline{}\newline{}Recurrent Neural Network Model\newline{}Graph Neural Network Model\newline{}Temporal Convolution Network Model & \ding{172}Intrinsic inductive biases tailored for causal sequential dependencies.\newline{}\ding{173}Moderate computational footprint suitable for deployment under tight edge resources.\newline{}\ding{174}Explicit structural modeling of inter-variable topological graphs when paired with graph convolutions. & \ding{172}Susceptible to gradient vanishing and memory bottlenecks over long temporal windows.\newline{}\ding{173}Constrained transferability compared to attention-based architectures, limiting cross-domain generalization. \\
    \midrule
    \rowcolor[rgb]{ .871,  .922,  .969} \ \newline{}\newline{}Prototype-based Model  & \ding{172}Explicitly characterizes normal manifolds using compact sets of learnable prototypes.\newline\ding{173}Modular framework easily integrated into diverse metric and representation backbones.  & \ding{172}Prototype representations require careful regularization to ensure geometric diversity and prevent representation collapse.\newline\ding{173}Scaling to large numbers of prototypes substantially increases optimization difficulty and inference latency.\\ 
    \midrule
    \rowcolor[rgb]{ .851,  .851,  .851} \ \newline{}\newline{}\newline{}{Classical Machine Learning}\newline{} & \ding{172}High operational transparency; tree ensembles provide intuitive decision pathways and feature importance rankings.\newline{}\ding{173}Well-suited for low-sample regimes with modest feature dimensionality. & \ding{172}Limited representational capacity for capturing complex non-linear temporal evolutions or cross-variable couplings.\newline{}\ding{173}Susceptible to the curse of dimensionality, leading to performance deterioration in high-channel settings.\newline{}\ding{174}Requires extensive manual feature engineering. \\
    \midrule
    \rowcolor[rgb]{ .851,  .851,  .851} \ \newline{}\newline{}{Statistical Method}\newline{} & \ding{172}Near-instantaneous execution speed, ideal for ultra-high-frequency online streaming.\newline{}\ding{173}Rigorous mathematical foundation providing exact probabilistic interpretation through analytical log-density.\newline{}\ding{174}Computationally scalable when formulated with structured marginal or low-rank joint covariance matrices. & \ding{172}Rigid parametric assumptions (such as Gaussianity and linearity) that frequently fail in real-world time series.\newline{}\ding{173}Incapable of modeling non-stationary drift and high-order temporal dynamics. \\
    \bottomrule
\end{tabular}}
    \label{tab:alg_ch2}
\end{table*}

\begin{table}
    \centering
    \caption{The algorithm selection suggestions under different selection criteria. }
    \includegraphics[width=0.8\linewidth]{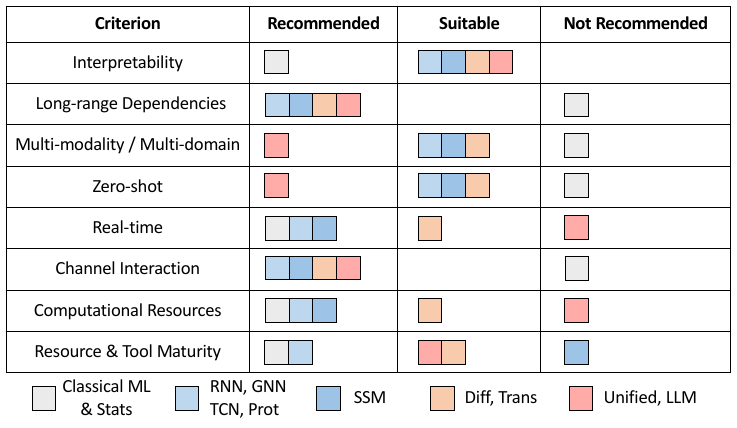}
    \label{tab:alg_ch3}
\end{table}

\subsection{Discussions on General Technologies}\label{sec:dis_gpt}
This subsection highlights core principles governing two widely used methodological strategies in TSAD: time series data augmentation and channel independence. A comprehensive literature taxonomy for both techniques is provided in SM~\ref{app:general_technologies}.

\textbf{Time Series Data Augmentation:} Data augmentation protocols, including noise injection, synthetic anomaly generation, and sample interpolation, serve to bolster representation robustness under self-supervised objectives \cite{oc_xuCalibratedOneclassClassification2022,AdaMemBLS,cont_shentuGeneralTimeSeries2024}. However, indiscriminate application of naive mixing techniques such as Mixup risks disrupting temporal causality and phase alignment. Practitioners should favor graph-guided or adversarially constrained perturbations that preserve underlying physical constraints while generating informative boundary-case nominal profiles.

\textbf{Channel Independence:} Decomposing multivariate sequences into isolated univariate channels significantly mitigates overfitting and enhances parameter sharing across heterogeneous datasets \cite{sv_nie2022time}. Nonetheless, because structural faults and channel-correlation anomalies fundamentally manifest through coordinated inter-sensor couplings, strict channel independence inevitably blinds detectors to multi-channel dynamics. Emerging architectures must harmonize univariate temporal modeling with adaptive cross-channel message passing, leveraging lightweight cross-attention or topological graph layers to dynamically capture synchronized inter-variable anomalies.
\section{Future and Opportunity}\label{sec:Future_and_Opportunity}
This section highlights critical bottlenecks in contemporary TSAD methodologies and outlines prospective directions for future research.

\subsection{Interpretability}\label{sec:Interpretability}
A persistent limitation in TSAD research is the disproportionate focus on benchmark detection accuracy at the expense of operational transparency \cite{sv_sarfraz2024positionquovadisunsupervised,sv_wuCurrentTimeSeries2022,TSB-AD}. In mission-critical deployments, interpretability is as essential as raw accuracy, as transparent decision boundaries allow domain operators to validate alarms and establish operational trust. Existing explainable TSAD models \cite{interpretability2,interpretability3,interpretability4} typically rely on heuristic attributions, such as ranking channel correlation coefficients or highlighting localized attention weights. However, such correlational signals fail to uncover underlying causal triggers, phase delay progressions, or physical feedback mechanisms \cite{XAI_IOT_gummadi2024xai,XAI_a_general_sipple2022}.

To provide actionable insights beyond correlational heuristics, meaningful TSAD interpretability must address three core dimensions:
\begin{itemize}
    \item \textbf{Temporal and Channel Specificity:} Detectors must accurately isolate the exact temporal boundaries of an anomaly while identifying the specific multivariate channels driving the anomalous departure.
    \item \textbf{Root Cause Analysis:} Beyond merely flagging anomalous occurrences, models should trace causal propagation pathways, identifying how initial variations in specific root-cause variables trigger downstream disturbances across interconnected channels over time.
    \item \textbf{Contextual Grounding:} Models should contextualize anomalous behaviors against nominal dynamics by isolating specific subsequences or shapelets responsible for the divergence, thereby allowing domain experts to inspect and verify the physical faults.
\end{itemize}

Establishing rigorous evaluation frameworks is equally vital for advancing explainable TSAD. Future progress depends on four methodological initiatives:
\begin{itemize}
    \item \textbf{Constructing Interpretable Datasets:} Controlled benchmark suites should incorporate verified ground-truth explanations. For instance, synthetic industrial datasets can simulate explicit multivariate interactions, such as coincident spikes in vibration and temperature, providing verifiable benchmarks for cross-channel attribution.
    \item \textbf{Structured Anomaly Annotation:} Benchmark annotations must advance beyond binary anomaly tags. Datasets should categorize anomaly taxonomies and explicitly link affected time points to their driving root causes, providing supervisory signals for explanatory models.
    \item \textbf{Developing a Formalized Language for Root Causes:} Comparing explainability across diverse models requires a structured descriptive grammar. This representation should systematically record anomaly categories, temporal boundaries, root-cause channels, key subsequences, and secondary cascading effects. Compiling such standardized descriptions into a unified evaluation repository will facilitate rigorous, cross-model benchmarking.
    \item \textbf{Advancing Explanatory Evaluation Metrics:} Quantitative criteria are needed to evaluate explanation fidelity, particularly explanation sparsity, defined as the minimal feature subset required to justify an alarm, and explanation stability, measured by attribution consistency under minor observational noise \cite{app_jacob2020exathlon,shapelet_ICLR,shapelet1}.
\end{itemize}

Integrating large language models also offers a promising direction for text-based interpretability \cite{sv_fonsEvaluatingLargeLanguage2024}. Although LLMs operate with black-box architectures, their natural language generation capabilities enable translating technical attribution maps and multivariate statistical discrepancies into coherent, operator-friendly summaries \cite{llm_LLMAD2024,llm_GPT4TS2023,llm_UNITS2024}. This capability can bridge the gap between complex model outputs and operational human decision-making.

\subsection{Zero-Shot Learning}
Zero-shot learning (ZSL) enables detectors to identify previously unseen anomaly patterns without requiring target-domain training data \cite{timemixer_PP_2025}. Driven by advancements in self-supervised pre-training, there is growing interest in building unified models that generalize across diverse time series domains \cite{cont_shentuGeneralTimeSeries2024,timemixer_PP_2025,llm_GPT4TS2023}. However, achieving reliable zero-shot detection in TSAD involves four primary challenges:

First, \textit{data scarcity and annotation bias} limit model training. True anomalies are inherently rare in practical operational data, and existing benchmarks rely almost exclusively on binary tags that lack granular descriptions of fault mechanisms. Injecting structured domain knowledge or synthetic perturbations offers a potential avenue to enrich training variety.
Second, \textit{distributional and behavioral divergence} complicates generalization. Anomalies manifest through varied characteristics across domains, ranging from subtle trend shifts in environmental monitoring to sudden transient spikes in industrial machinery. A model trained on a specific set of dynamics often fails when deployed in an unfamiliar regime.
Third, \textit{structural heterogeneity} poses architectural constraints. Real-world applications feature sequences of arbitrary lengths, irregular sampling rates, and variable channel dimensions \cite{llm_UNITS2024}. Detectors with rigid input layers cannot process these heterogeneous inputs without lossy interpolation or channel selection.
Fourth, a clear \textit{performance gap remains between few-shot adaptation and true zero-shot inference}. Current large-scale time series models frequently require fine-tuning on a small fraction of domain-specific samples (such as 5\% of the target dataset) to yield acceptable performance \cite{llm_UNITS2024,llm_GPT4TS2023}. 

Closing this gap is essential for practical deployment across diverse industries. A truly capable zero-shot detector should accurately identify anomalies upon initial deployment, avoiding the high costs and operational latency associated with target-specific data collection and retraining.
\subsection{Multi-Modality Learning}
Classical machine learning and deep learning TSAD methods typically focus solely on sequential data. In contrast, LLM-based unified models often incorporate both sequential and textual modalities. One common approach involves projecting sequential data into vector embeddings and feeding them into LLMs. To align these embedding spaces, researchers usually redesign the input layer and fine-tune it on downstream time series datasets. An alternative approach uses prompt learning to format sequential data as text prompts, although directly representing numerical values as text often degrades their intrinsic mathematical semantics.

In parallel, converting time series data into visual representations allows models to leverage the rich priors embedded in pre-trained vision-language models \cite{sv_zhangLargeLanguageModels2024}. For example, Insight Miner \cite{dis_moon2023anymal} fine-tunes LLaVA to translate windowed time series images into descriptive language. Similarly, CLIP-LSTM converts stock market data into textual and visual representations, utilizing CLIP features for downstream tasks. Despite these early attempts, research on multi-modal TSAD remains scarce. Given the growing emphasis on multi-modal perception across AI domains, effectively integrating diverse modalities and aligning their embedding spaces represents a critical frontier for TSAD.

Despite these developments, current multimodal TSAD frameworks remain largely limited to shallow feature concatenation \cite{sv_zhangLargeLanguageModels2024,dis_moon2023anymal,llm_UNITS2024}. In these setups, auxiliary modalities are merely appended as side inputs, without deep cross-modal reasoning. To establish a more tightly integrated multimodal framework, future research should explore three key opportunities:

First, \textit{integrating contextual metadata throughout training}: Time series signals in complex engineering systems are accompanied by equipment maintenance logs, sensor topology descriptions, and operational manuals. Incorporating these textual sources directly into the representation space enables models to distinguish benign operating adjustments from genuine equipment faults.
Second, \textit{multiround conversational feedback}: To facilitate collaborative diagnostics, TSAD systems require an interactive loop where human feedback refines model outputs. When operators correct a false alarm through natural language, the system should dynamically update its confidence scores and adjust future predictions.
Third, \textit{interactive multimodal prompting}: Practical diagnostics benefit from flexible operator inputs, combining natural language queries with visual annotations (such as bounding boxes on sensor charts) and temporal window selections. Compiling visual prompts into spatial attention masks and text instructions into logical constraints provides an intuitive interface for root cause investigation.
\subsection{Multi-Domain Learning}
Unlike conventional domain-specific models that require separate training for data from each domain, a unified TSAD model should share parameters across multiple domains and handle various tasks simultaneously. It should exhibit key characteristics including: (1) structural agnosticism, allowing the model to process time series data of different lengths and variables without modifying its architecture, and (2) parameter sharing, enabling the model to utilize shared representations for concurrent task handling \cite{llm_UNITS2024}.

Considering previous discussions, we anticipate that future unified TSAD models need the following characteristics:
1) Zero-shot capability: Current large-scale time series models often require fine-tuning with few-shot data when switching scenarios. Making TSAD models commercially viable will depend on enabling them to perform anomaly detection without any fine-tuning.
2) Dynamic threshold selection: Given the inherent differences in data across domains, the model should adaptively select thresholds based on the current scenario. As discussed, using dynamic thresholds is essential for the current model design. For TSAD models that learn across multiple domains, it becomes even more important—not just to handle fixed thresholds failing over time, but also to deal with the added uncertainty of thresholds varying across domains.
3) Interpretability: Identifying anomalies in time series data is only the initial step. In practical applications, early alerts are essential, but the model must also explain the anomalies to facilitate subsequent tasks, such as identifying root causes and suggesting fixes.
\subsection{Foundational Models}
The emergence of foundational models presents a transformative opportunity for time series anomaly detection, offering potential to handle diverse operational tasks within a unified architecture. Current research directions generally fall into two categories: native time series foundational models and adapted LLM-based architectures.

Native time series foundational models are pre-trained directly on extensive collections of time series data to learn cross-domain temporal dynamics (as detailed in SM~\ref{sec:Unified_TSAD}). A common implementation strategy decomposes multivariate sequences into independent univariate channels, which facilitates zero-shot transfer across datasets with varying channel counts. However, the limited availability of high-quality, open-source time series archives constrains the scale and generalization capabilities of these models compared to their vision and language counterparts. Additionally, relying solely on numerical sequences prevents these models from leveraging external contextual knowledge across domains.

Adapted LLM-based models explore transferring linguistic representations to sequential numerical inputs (as discussed in SMs~\ref{sec:pred_llm} and \ref{sec:exp_llm}). While pre-trained LLMs offer strong contextual reasoning and text generation capabilities, they are not inherently designed for continuous numerical signals. Standard text tokenizers are optimized for discrete linguistic tokens, often struggling to capture high-frequency oscillations, exact numerical magnitudes, and temporal phase relationships without precision loss. 

Looking forward, a promising trajectory lies in developing unified multimodal foundational models that jointly process numerical time series, structural system topologies, and textual documentation. Such architectures can integrate the numeric precision of native time series encoders with the contextual reasoning of language backbones, providing a robust and generalizable foundation for next-generation anomaly detection.
\section{Conclusion}\label{sec:Conclusion}
This survey provides a comprehensive review of advances in time series anomaly detection by establishing a rigorous taxonomy, categorizing modern algorithmic paradigms, and conducting an empirical evaluation of benchmark protocols. By refining anomaly definitions and methodology classifications, this work establishes a structured taxonomy aligned with contemporary research. We evaluate representative approaches, analyzing their operational strengths, computational constraints, and practical suitability while offering concrete algorithmic selection criteria. Additionally, this survey examines modern architectural designs introduced to time series anomaly detection, including diffusion formulations, state space models, large language models, and unified cross-domain architectures. In addition, we categorize evaluation protocols and present empirical analyses addressing latency, stability, and evaluation bias. Finally, we delineate persistent challenges and open frontiers across causal interpretability, zero-shot transfer, multimodal integration, adaptive thresholding, and foundational models, offering a coherent roadmap for future research and operational deployment.

\bibliographystyle{IEEEtranN}
\bibliography{sn-bibliography}

\begin{IEEEbiography}
   [{\includegraphics[width=1in,height=1.25in,clip,keepaspectratio]{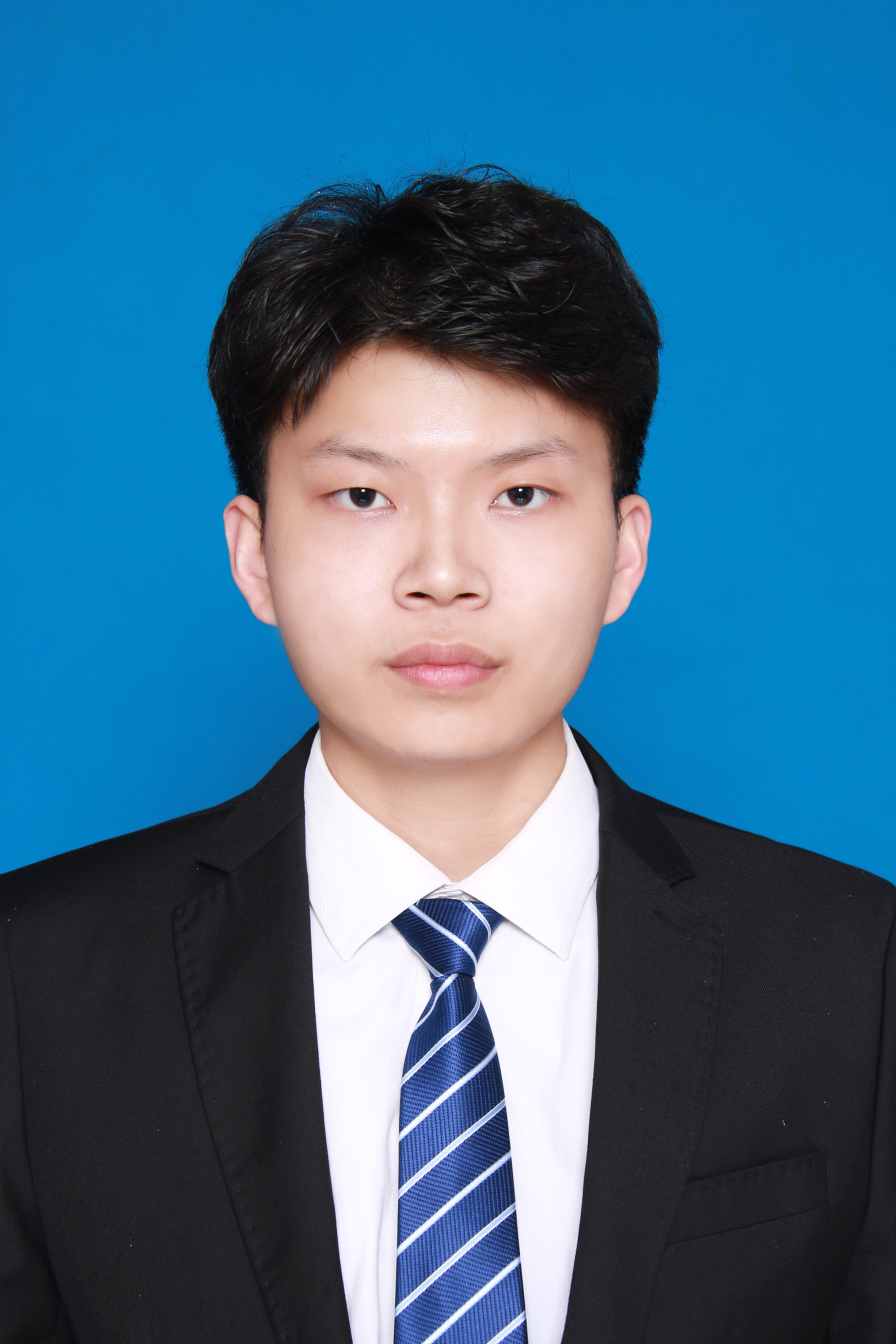}}]
{Zhijie Zhong} received the B.S. degree in 2022 from the Harbin Engineering University, Harbin, China and he is currently pursuing the Ph.D. degree in the School of Future Technology, South China University of Technology, Guangzhou, China. His research interests include data mining, machine learning, time series analysis, anomaly detection, and large language model (LLM).
\end{IEEEbiography}

\begin{IEEEbiography}
[{\includegraphics[width=1in,height=1.25in,clip,keepaspectratio]{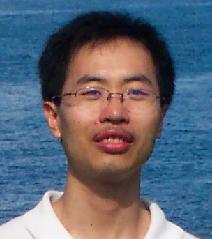}}]
{Zhiwen Yu} received the Ph.D. degree from the City University of Hong Kong, Hong Kong, in 2008. He is a Professor with the School of Computer Science and Engineering, South China University of Technology, China. He has authored or co-authored more than 200 refereed journal articles and international conference papers, including more than 90 articles in the journals of IEEE Transactions, with an H-index of 56 and Google citations of 16000. He is an Associate Editor of IEEE Transactions on Systems, Man, and Cybernetics: Systems and IEEE Transactions on Neural Networks and Learning Systems. Ranked among the top 2\% of scientists worldwide in Stanford University’s 2024 World’s Top 2\% Scientists List.
\end{IEEEbiography}

\begin{IEEEbiography}
     [{\includegraphics[width=1in,height=1.25in,clip,keepaspectratio]{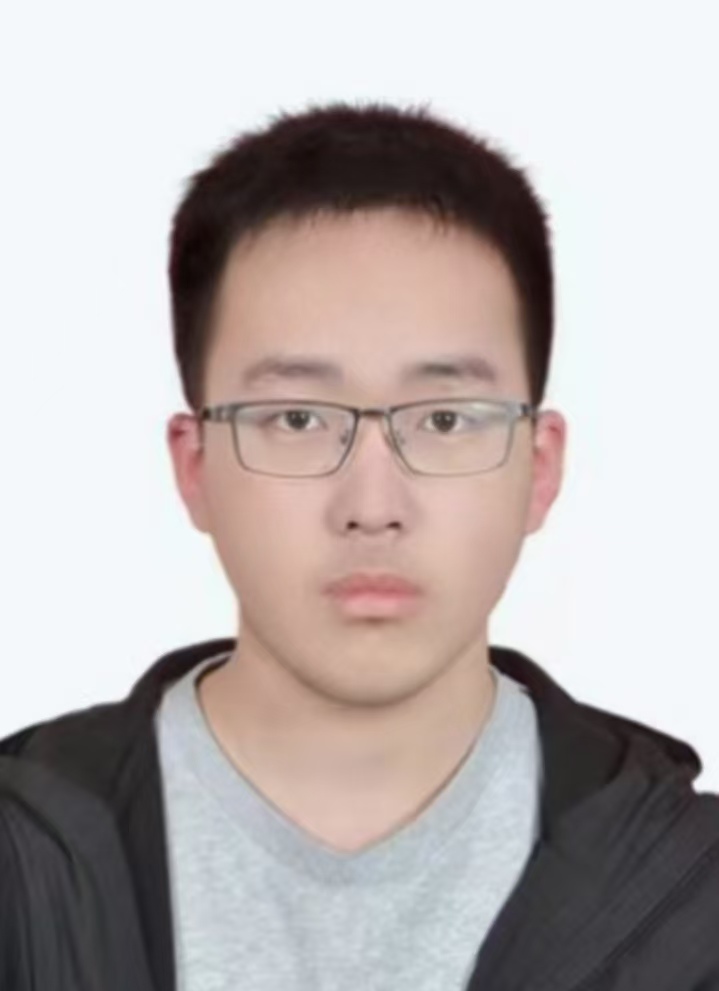}}]
{Pengyu Li} received the B.S. degree in 2024 from Nanchang University, Jiangxi, China, and he is currently pursuing the M.S. degree at South China University of Technology, Guangzhou, China. His research interests include machine learning, time series anomaly detection, and large language models (LLMs).
\end{IEEEbiography}

\begin{IEEEbiography}[{\includegraphics[width=1in,height=1.25in,clip,keepaspectratio]{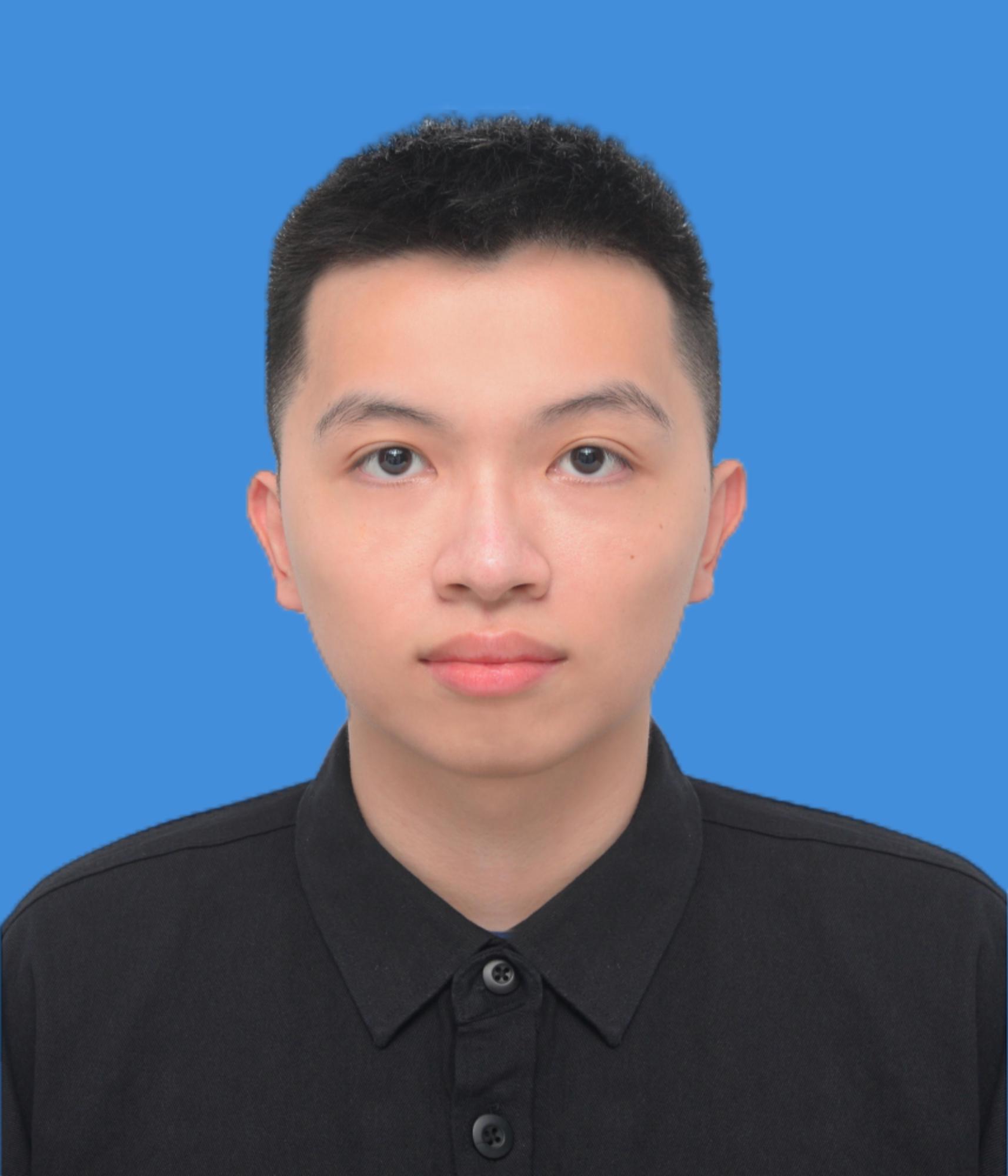}}]
{Liujian Zhang} received the B.S. degree from the School of Computer Science and Engineering, South China University of Technology, Guangzhou, China in 2022. He is currently pursuing the Ph.D. degree in the School of Future Technology, South China University of Technology, Guangzhou, China. His research interests include machine learning and transfer learning.
\end{IEEEbiography}

\begin{IEEEbiographynophoto}
    {Qiaoli Ni} is with HUAWEI Technologies Co., Ltd., Shenzhen, Guangdong 518066, China.
\end{IEEEbiographynophoto}

\begin{IEEEbiography}
    [{\includegraphics[width=1in,height=1.25in,clip,keepaspectratio]{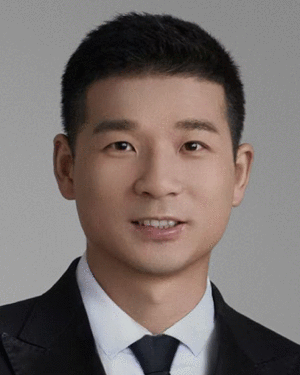}}]
    {Jun Jiang} is currently a post-doctoral researcher with the Department of New Networks, Pengcheng Laboratory, Shenzhen, China. He received his Ph.D. degree in computer science and technology from the South China University of Technology, Guangzhou, China, in 2022. He was a visiting student at Pengcheng Laboratory, Shenzhen, China, from March 2022 to April 2023. He worked as a lecturer with the College of Information Science and Technology, Nanjing Forestry University, Nanjing, China, from 2023 to 2024. His research interests include data stream classification, anomaly detection, cloud computing, and the Internet of Things.
\end{IEEEbiography}

\begin{IEEEbiography}
    [{\includegraphics[width=1in,height=1.25in,clip,keepaspectratio]{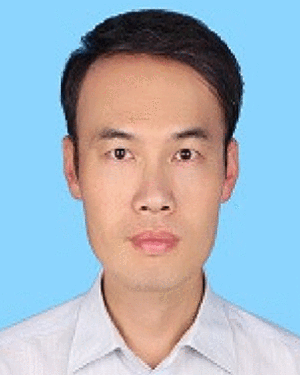}}]
    {Jianming Lv} received the B.S. degree in computer science from Sun Yat-sen University, Guangzhou, China, in 2002, and the Ph.D. degree from the Institute of Computing Technology, University of Chinese Academy of Sciences, Beijing, China, in 2008. He is currently an Associate Professor with the South China University of Technology, Guangzhou, China. He has authored or coauthored more than 30 papers with prestigious venues, including IEEE Transactions on Pattern Analysis and Machine Intelligence, the IEEE Transactions on Services Computing, ACM TOMMCAP, Computer Networks, CVPR, and CIKM. His research interests include data mining, computer vision, and distributed computing and privacy. He is a member the ACM and the CCF.
\end{IEEEbiography}

\begin{IEEEbiography}
     [{\includegraphics[width=1in,height=1.25in,clip,keepaspectratio]{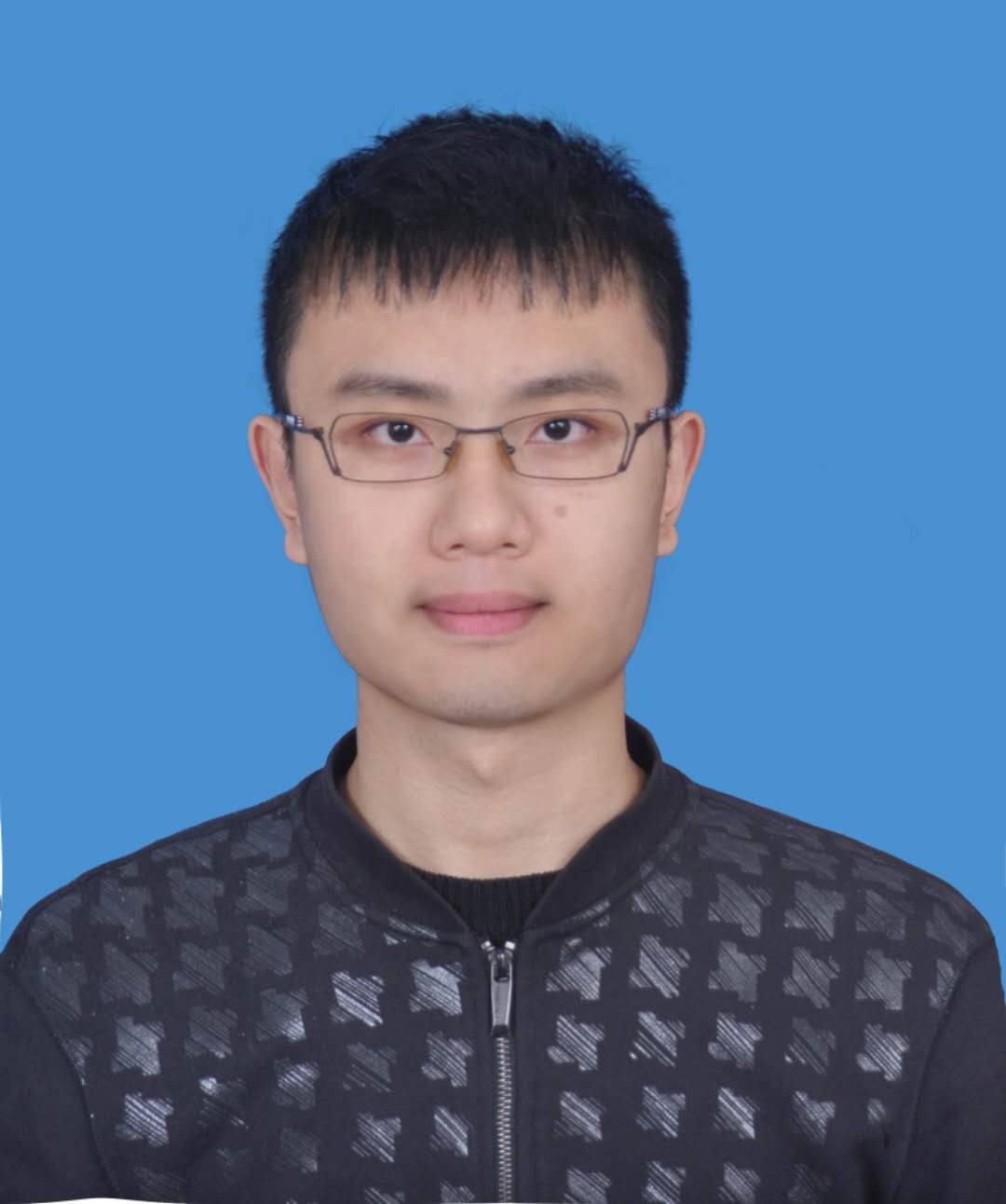}}]
{Kaixiang Yang (M'21)} received the B.S. degree from the University of Electronic Science and Technology of China, Chengdu, China, in 2012, the M.S. degree from the Harbin Institute of Technology, Harbin, China, in 2015, and the Ph.D. degree from the School of Computer Science and Engineering, South China University of Technology, Guangzhou, China, in 2020. He was a Research Engineer with 7th Research Institute, China Electronics Technology Group Corporation, Guangzhou, China, from 2015 to 2017 and Postdoctoral Researcher with Zhejiang University from 2021 to 2023. He is currently an Associate Professor with the School of Computer Science and Engineering, South China University of Technology. His research interests include pattern recognition, machine learning, and industrial data intelligence. He is an Associate Editor for Pattern Recognition. He is a senior member of China Computer Federation
\end{IEEEbiography}

\begin{IEEEbiography}
     [{\includegraphics[width=1in,height=1.25in,clip,keepaspectratio]{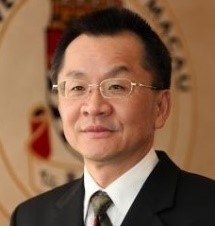}}]
{C. L. Philip Chen (S'88-M'88-SM'94-F'07)}
graduated from the University of Michigan at Ann Arbor, Ann Arbor, MI, USA, in 1985, and the Ph.D. degree in electrical engineering from Purdue University, West Lafaette, Indiana, USA, in 1988.,He is currently the Chair Professor and the Dean of the College of Computer Science and Engineering, South China University of Technology, Guangzhou, China. Being a Program Evaluator of the Accreditation Board of Engineering and Technology Education in the USA, for computer engineering, electrical engineering, and software engineering programs, he successfully architects the University of Macau’s Engineering and Computer Science programs receiving accreditations from Washington/Seoul Accord through the Hong Kong Institute of Engineers, of which is considered as his utmost contribution in engineering/computer science education for Macau as the former Dean of the Faculty of Science and Technology. His current research interests include cybernetics, systems, and computational intelligence.,Prof. Chen is a Fellow of the American Association for the Advancement of Science, the International Association for Pattern Recognition, the Chinese Association of Automation (CAA), and the Hong Kong Institution of Engineers. He is a member of Academia Europaea, the European Academy of Sciences and Arts, and the International Academy of Systems and Cybernetics Science. He received the IEEE Norbert Wiener Award in 2018 for his contribution in systems and cybernetics, and machine learnings. He was also a highly cited researcher by Clarivate Analytics in 2018 and 2019. He was a recipient of the 2016 Outstanding Electrical and Computer Engineers Award from his Alma Mater, Purdue University, West Lafayette, IN, USA, in 1988. He was the President of the IEEE Systems, Man, and Cybernetics Society from 2012 to 2013. He was the Editor-in-Chief for IEEE Transactions on Systems, Man, and Cybernetics: Systems from 2014 to 2019 and IEEE Transactions on Cybernetics from 2019 to 2021). He was an Associate Editor for IEEE Transactions on Artificial Intelligence and IEEE Transactions on Fuzzy Systems. He was the Chair of TC 9.1 Economic and Business Systems of the International Federation of Automatic Control from 2015 to 2017. He is a Vice-President of the CAA.
\end{IEEEbiography}

\end{document}